\documentclass{ieeeaccess}
\usepackage[x11names,table]{xcolor}
\usepackage{cite}
\usepackage{amsmath,amssymb,amsfonts}
\usepackage{algorithmic}
\usepackage{graphicx}
\usepackage{textcomp}

\usepackage{multirow}%
\usepackage{amsmath,amssymb,amsfonts}%
\usepackage{booktabs}%

\usepackage{makecell}  
\usepackage{enumitem}

\def\BibTeX{{\rm B\kern-.05em{\sc i\kern-.025em b}\kern-.08em
    T\kern-.1667em\lower.7ex\hbox{E}\kern-.125emX}}
\begin{document}
\history{Date of publication xxxx 00, 0000, date of current version xxxx 00, 0000.}
\doi{10.1109/ACCESS.2017.DOI}

\title{Onboard Optimization and Learning: A Survey}

\author{\uppercase{Monirul Islam Pavel}\authorrefmark{1}, 
\uppercase{Siyi Hu\authorrefmark{2}, Mahardhika Pratama \authorrefmark{1} \IEEEmembership{Senior Member, IEEE}},
\uppercase{Ryszard Kowalczyk}\authorrefmark{1,3}}

\address[1]{School of Computer Science and Information Technology, Adelaide University, Australia}
\address[2]{School of Electrical Engineering, Computing and Mathematical Sciences, Curtin University, Australia}

\address[3]{Systems Research Institute, Polish Academy of Sciences, Warsaw, Poland}

\markboth
{M.I. Pavel \headeretal: Onboard Optimization and Learning: A Survey}
{M.I. Pavel \headeretal: Onboard Optimization and Learning: A Survey}
\corresp{Corresponding author: Siyi Hu (e-mail:siyi.hu@curtin.edu.au).}

\begin{abstract}
Onboard learning is a transformative approach in edge AI, enabling real-time data processing, decision-making, and adaptive model training directly on resource-constrained devices without relying on centralized servers. This paradigm is crucial for applications demanding low latency, enhanced privacy, and energy efficiency. However, onboard learning faces challenges such as limited computational resources, high inference costs, and security vulnerabilities.
This survey explores a comprehensive range of methodologies that address these challenges, focusing on techniques that optimize model efficiency, accelerate inference, and support collaborative learning across distributed devices. Approaches for reducing model complexity, improving inference speed, and ensuring privacy-preserving computation are examined alongside emerging strategies that enhance scalability and adaptability in dynamic environments. By bridging advancements in hardware-software co-design, model compression, and decentralized learning, this survey provides insights into the current state of onboard learning to enable robust, efficient, and secure AI deployment at the edge.
\end{abstract}

\begin{keywords}
edge AI, edge computing, inference,  model compression, on-device learning, onboard training, optimization
\end{keywords}

\titlepgskip=-15pt

\maketitle

\section{Introduction}
\label{sec:introduction}
\PARstart{T}{he} increasing adoption of edge artificial intelligence (AI) has driven a paradigm shift toward onboard optimization and learning, enabling devices to process data, make decisions, and update models locally without relying on centralized servers~\cite{dhar2021survey}. Traditional cloud-based AI architectures face significant limitations, including high latency, bandwidth constraints, and security risks associated with data transmission~\cite{bourechak2023confluence, zhu2023low}. These issues are particularly critical for applications such as autonomous vehicles, industrial IoT, and smart infrastructure, where real-time decision-making and responsiveness are essential. Cloud inference often suffers from 200 to 300 ms delays under constrained uplink bandwidth, in contrast to edge deployments that maintain six times lower latency \cite{li2019edge}. Such disparities are required in Intelligent Transportation Systems (ITS), where applications like platooning, cooperative perception, and road-safety signaling demand 10 to 100 ms end-to-end responsiveness, requirements that centralized cloud-based AI deployments cannot consistently satisfy \cite{arthurs2021taxonomy}. Onboard learning addresses these challenges by performing computations directly on resource-constrained devices, ensuring faster, more efficient, and privacy-preserving AI inference and adaptation~\cite{cai2020tinytl, hoefler2021sparsity, imteaj2021survey}.

\begin{table*}[t]
\centering
\caption{Taxonomy of key techniques and challenges. (H = helps, N = neutral/mixed, U = hurts).}
\small
\renewcommand{\arraystretch}{1.7}
\resizebox{\textwidth}{!}{
\begin{tabular}{l l c c c c c c c c c c l}
\toprule
\multirow{3}{*}{\makecell[l]{\textbf{Topic}}} &
\multirow{3}{*}{\textbf{Techniques}} &
\multicolumn{3}{c}{\textbf{Performance Constraint}} &
\multicolumn{2}{c}{\textbf{Latency \& Comm}} &
\multicolumn{2}{c}{\textbf{Hardware Adaptability}} &
\multicolumn{3}{c}{\textbf{Optimization Complexity}} &
\multirow{3}{*}{\makecell[l]{\textbf{Summary}}} \\
\cmidrule(lr){3-5} \cmidrule(lr){6-7} \cmidrule(lr){8-9} \cmidrule(lr){10-12}
& & HCC & MSC & HPC & LI & CO & HC & HEE & TDB & PT & OC & \\
\midrule

\multirow{5}{*}{\makecell[l]{\textbf{Model}\\\textbf{Compression}}}
& Pruning                & H & H & H & H & N & H & N & N & U & N & Structured helps latency, unstructured needs sparse support. \\
& Quantization           & H & H & H & H & N & H & N & N & N & N & INT8 helps on CPUs/NPUs, mixed on older GPUs. \\
& Knowledge Distillation & H & H & N & H & N & H & N & N & N & N & Smaller student preserves accuracy under shift. \\
& Neural Architecture Search    & H & H & N & H & N & H & H & U & U & U & Weight-sharing reduces cost, full DARTS heavy. \\
& Adapter Based Fine-Tuning & H & H & N & H & N & H & H & N & N & N & Updating only compact adapter layers, reducing retraining cost. \\
\midrule

\multirow{3}{*}{\makecell[l]{\textbf{Efficient}\\\textbf{Inference}}}
& Computation Offloading & N & N & N & H & U & N & H & N & N & N & Helps with stable link, hurts with poor RTT/privacy. \\
& Model Partitioning     & H & N & N & H & N & H & H & N & N & N & Effective if partitions balance tensor/activation sizes. \\
& Early Exit Strategies  & H & N & H & H & N & N & H & N & N & N & Gains when gating calibrated, minor mem overhead. \\
\midrule

\multirow{3}{*}{\makecell[l]{\textbf{Decentralized}\\\textbf{Learning}}}
& Federated Learning     & N & N & N & N & H & N & H & U & U & N & Comm saved vs. central, accuracy may drop non-IID. \\
& Split Learning         & H & H & H & U & U & N & H & U & U & U & Distributed onboard training partitioned between edge and coordinated nodes.\\
& Continual Learning + DL& N & N & N & N & N & N & H & N & N & N & Replay/regularization mitigate drift. \\
& Adaptive Learning + DL & H & H & N & N & H & N & N & N & N & N & Lightweight adapters handle task shift. \\
\midrule

\multirow{3}{*}{\makecell[l]{\textbf{Security}\\\textbf{\& Privacy}}}
& Privacy Protection     & H & H & U & N & N & N & H & N & N & N & DP helps privacy, high noise may hurt accuracy. \\
& Secure Model Execution & H & H & U & N & N & H & H & N & N & N & TEEs/attestation, overhead depends on enclave I/O. \\
& Explainable AI         & N & N & N & N & N & N & H & N & N & N & Transparency/compliance, minimal performance impact. \\
\midrule

\multirow{3}{*}{\makecell[l]{\textbf{Advanced}\\\textbf{Topics}}}
& Compression + CL       & N & N & N & N & N & H & N & N & N & N & Combine pruning/quantization with replay/adapters. \\
& Scalability            & H & H & H & H & H & H & H & H & H & H & Fleet-level deployment patterns. \\
& Hardware Co-Design     & H & H & H & H & H & H & H & H & H & H & Joint model-hardware tuning. \\
\bottomrule
\end{tabular}
}
\label{tab:taxonomy}
\vspace{0.5ex}

{\footnotesize
\begin{tabular}{@{}p{\textwidth}@{}}
    \textbf{Performance Constraints:} High Computational Cost (HCC), Memory \& Storage Constraints (MSC), High Power Consumption (HPC) \\
    \textbf{Latency and Communication:} Latency Issues (LI), Communication Overhead (CO) \\
    \textbf{Hardware Adaptability:} Hardware Compatibility (HC), Heterogeneous Edge Environments (HEE) \\
    \textbf{Optimization Complexity:} Training \& Deployment Bottlenecks (TDB), Performance Trade-off (PT), Optimization Complexity (OC)
\end{tabular}
}

\end{table*}

Despite its advantages, onboard learning introduces several challenges. Edge devices typically have limited computational power, memory, and energy resources, making it difficult to deploy and run complex AI models efficiently~\cite{wang2020convergence, deng2020edge}. Furthermore, optimizing inference for real-time applications requires balancing model accuracy, latency, and power consumption. Additionally, decentralized learning frameworks must facilitate model training and updates across multiple devices without compromising privacy or requiring frequent cloud synchronization~\cite{bourechak2023confluence, zhu2023low}. Finally, ensuring security and robustness against adversarial threats is essential for onboard AI systems operating in dynamic and often untrusted environments.

This survey systematically categorizes onboard learning methodologies into five fundamental aspects. The first aspect, \textit{Model Compression}, focuses on reducing the size and complexity of AI models to fit within the computational constraints of edge devices. Techniques such as pruning, quantization, knowledge distillation, and neural architecture search are explored to optimize model efficiency without significantly degrading performance~\cite{hoefler2021sparsity, imteaj2021survey}.  
The second aspect, \textit{Efficient Inference}, examines strategies that minimize latency and energy consumption while maintaining real-time AI responsiveness. Methods such as model partitioning, early exit mechanisms, and computation offloading are discussed to improve inference efficiency~\cite{wang2020convergence, deng2020edge, lin2022device, zhu2024device}.  
The third aspect, \textit{Decentralized Learning}, enables collaborative training and model adaptation across multiple devices without relying on a centralized server. Federated learning, continual learning (CL), and adaptive techniques are analyzed to improve scalability and maintain privacy while adapting to dynamic data distributions~\cite{bourechak2023confluence, zhu2023low, cai2020tinytl, hoefler2021sparsity}.  
The fourth aspect, \textit{Security and Privacy}, addresses the risks associated with deploying AI models in untrusted environments. Methods such as differential privacy, adversarial robustness, and encrypted model updates are discussed to safeguard onboard AI systems against potential threats~\cite{wang2020convergence, zhu2023low}. 
The final aspect \textit{Advanced Topics} discusses research directions that enhance the effectiveness, scalability, and resilience of optimized onboard learning. This includes hardware-software co-design to optimize AI, bridging model compression with CL to increase flexibility while preserving efficiency, scalability, and standardization for efficient deployment across heterogeneous edge platforms ~\cite{luo2022lightnas, chitty2022neural, dhar2021survey}. 

The summarization of this survey paper’s taxonomy is presented in Table~\ref{tab:taxonomy}, which provides a synthesized overview of how each aspect is analyzed throughout this review. The taxonomy is organized across four critical system dimensions, including performance constraints, latency \& communication, and hardware adaptability, each representing distinct challenge in onboard AI deployment for optimized learning. The table employs qualitative indicators: {H (Helps), N (Neutral/Mix), and U (Hurts)}, to denote whether the discussed aspects and techniques are generally beneficial, have limited or context-dependent influence, or tend to introduce additional system-level burdens, respectively. By structuring onboard learning around these core aspects, this survey provides a comprehensive review of the methodologies optimizing AI for edge computing. The remainder of this paper is organized as follows. Section~\ref{sec:model_compression} explores techniques for model compression and efficiency. Section~\ref{sec:efficient_inference} discusses inference optimization strategies. Section~\ref{sec:decentralized_learning} covers decentralized and adaptive learning methods. Section~\ref{sec:security_privacy} examines security and privacy considerations. Section ~\ref{advanced_topics} presents emerging advanced topics for optimized onboard learning. The survey concludes with an overview and discussion of future directions for onboard AI \footnote{Terms such as on-device learning, onboard learning, and edge-based learning or edge AI are often used interchangeably in the existing literature. In this paper, we will use 'onboard learning' for standardization.}.

\textbf{Contributions.}
Compared with existing surveys on edge AI, model compression, and federated learning, this work adopts an explicitly onboard-centric and cross-layer perspective. Its main contributions can be summarized as follows.  
\begin{itemize}
    \item Structured review of onboard and edge-native learning techniques across model compression, inference optimization, decentralized and continual learning, security, and deployment frameworks, with emphasis on operation under strict resource constraints.
    \item Established unified view of how constraints on computation, memory, energy, and communication delimit the feasible design space of optimized onboard learning systems.
 latency, communication, and hardware heterogeneity.
    \item Synthesized interdependency and recurring quantitative trends to show how state-of-the-art techniques interact in realistic onboard pipelines for system-level insight rather than a purely descriptive enumeration of methods.
\end{itemize}

\textbf{Review Scope and Selection.}
This survey systematically investigates how AI models are being optimized, trained, and deployed on resource-constrained edge devices, addressing the research question, \textit{"What techniques enable optimized onboard learning while maintaining model performance under hardware constraints?"} Following the PRISMA flow diagram for literature inclusion (shown in Figure\ref{fig_0}), We searched WoS, IEEE Xplore, ACM Digital Library, ArXiv, and SpringerLink for works published between 2016 to 2025. 
The initial query retrieved 970 papers. After screening for quality (preferring Q1-level journals and CORE-ranked conferences) and relevance to onboard (on-device/edge) AI, 
283 papers were retained for analysis. Overall, the literature searching criteria and queries are broadly described below:

\Figure[t!](topskip=0pt, botskip=0pt, midskip=0pt)[width=0.47\textwidth]{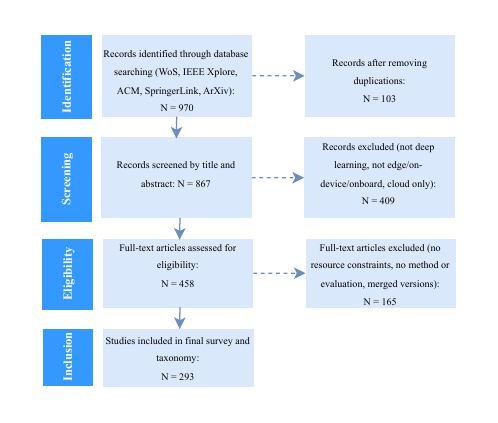}
{%
  \begin{minipage}[t]{0.4\textwidth} % <-- [t] makes the box top-aligned
    \raggedright
    \textbf{PRISMA flow diagram of the identification process for the sample of articles.}
  \end{minipage}%
  \label{fig_0}%
}

\begin{itemize}[leftmargin=*,noitemsep,topsep=0pt]
    \item {\small
    (Onboard $\,||\,$ On-device $\,||\,$ Edge) 
    $\,\&\,$ (Inference $\,\&\,$ 
    (offloading $\,||\,$ Model partitioning $\,||\,$ 
    Right-sizing $\,||\,$ Accelerator))
    }
    
    \item {\small
    (Onboard $\,||\,$ On-device $\,||\,$ Edge) 
    $\,\&\,$ (Optimization $\,\&\,$ 
    (Quantization $\,||\,$ Pruning $\,||\,$ Knowledge Distillation$\,||\,$ NAS$\,||\,$ Inference))
    }
    \item {\small
    (Onboard $\,||\,$ On-device $\,||\,$ Edge) 
    $\,\&\,$ (Training $\,\&\,$ 
    (Privacy $\,||\,$ Decentralization $\,||\,$ 
    Explainable $\,||\,$ Dimensionality reduction))
    }
    \item {\small
    (Onboard $\,||\,$ On-device $\,||\,$ Edge) 
    $\,\&\,$ (Learning $\,\&\,$ 
    (Online learning $\,||\,$ Federated learning $\,||\,$ 
    Continual learning))
    }
\end{itemize}

% \textbf{Identification}
% \begin{itemize}[leftmargin=*]
%     \item Records identified through database searching (WoS, IEEE Xplore, ACM DL, SpringerLink): $n = 970$
%     \item Records after duplicate removal: $n = N_{\text{id\_dedup}}$
% \end{itemize}

% \textbf{Screening}
% \begin{itemize}[leftmargin=*]
%     \item Records screened by title and abstract: $n = N_{\text{id\_dedup}}$
%     \item Records excluded (not deep learning, not edge/on-device/onboard, cloud only): $n = N_{\text{screen\_excluded}}$
% \end{itemize}

% \textbf{Eligibility}
% \begin{itemize}[leftmargin=*]
%     \item Full text articles assessed for eligibility: $n = N_{\text{full\_assessed}}$
%     \item Full text articles excluded (no resource constraints, no method or evaluation, merged versions): $n = N_{\text{full\_excluded}}$
% \end{itemize}

% \textbf{Included}
% \begin{itemize}[leftmargin=*]
%     \item Studies included in final survey and taxonomy: $n = 259$
% \end{itemize}

\section{Model Compression} \label{sec:model_compression}

\begin{table*}
  \caption{Summary of model compression methods for optimized onboard learning.}
  \small
  \renewcommand{\arraystretch}{1}
  \centering
  \resizebox{\textwidth}{!}{ 
  \begin{tabular}{lll p{6.5cm}}
    \toprule
    \textbf{Technique} & \textbf{Category} & \textbf{Reference} & \textbf{Key Highlights} \\
    \midrule
    \multirow{4}{*}{\textbf{Pruning}}
      & Unstructured vs. Structured & \cite{xia2022structured, yu2020easiedge, ro2021autolr,li2022optimizing, kong2023towards, zhang2025optimization,yu2024edge,joardar2022realprune,yin2024workload} & Sparse vs. hardware-friendly pruning.\\
      & Pruning at Initialization & \cite{tanaka2020pruning, frantar2023sparsegpt, kohama2023single, chang2023iterative} & Early pruning for efficiency.\\
      & Pruning During Training & \cite{jiang2023computation, guo2024online, jelvcicova2022delta} & Dynamic real-time pruning.\\
      & Post-Training Pruning & \cite{li2025improving,lazarevich2021post,kwon2022fast,xu2025lightweight} & Deduct redundant weight after full model training, without retraining through minimal calibration.\\
    \midrule
    \multirow{4}{*}{\textbf{Quantization}}
      & Post-Training Quantization & \cite{jeon2022mr, kim2024towards} & Converts FP32 to INT8, faster but less accurate.\\
      & Quantization-Aware Training & \cite{rasch2023hardware, shen2021once,zhang2025survey,mahmudov2025quantedge} & Simulates quantization during training, more accurate but slower.\\
      & Mixed Precision Quantization & \cite{koryakovskiy2023one, guo2020single, dong2019hawq, yao2021hawq, wang2019haq, zhao2025hardware, van2020bayesian} & Variable bit-width layers.\\
      & Hybrid Run-Time Quantization & \cite{zheng2025fedhq, kim2025oaken, zeng2025subkv} & PTQ calibration with runtime adaptability.\\
    \midrule
    \multirow{3}{*}{\textbf{Knowledge Distillation}}
      & Teacher-Student Learning & \cite{gou2021knowledge, wang2021knowledge} & Model compression via knowledge transfer.\\
      & KD for Edge Devices & \cite{sepahvand2022teacher, crowley2018moonshine, xiao2024knowledge} & Reduces FLOPs and memory.\\
      & Federated KD & \cite{itahara2021distillation, qu2020quantization, luo2022keepedge, he2022learning, jin2022personalized, qi2022fedbkd} & Distills global to local models.\\
    \midrule
    \multirow{3}{*}{\textbf{Neural Architecture Search}}
      & Hardware-Aware NAS & \cite{lyu2021resource, lee2021hardware, zhou2024hgnas} & Auto-optimized for edge devices.\\
      & NAS Compression Challenges & \cite{liu2022survey, liu2018darts} & High search cost, memory use.\\
      & Few-Shot NAS & \cite{elsken2020meta, kim2023device} & Fast, memory-efficient architecture search.\\
    \midrule
    \multirow{3}{*}{\textbf{Adapter Based Fine-Tuning}}
      & Low-Rank and Quantized Adaptation & \cite{hu2022lora, dettmers2023qlora, guo2023lq, jeon2024l4q,bondarenko2024low} & Efficient updates using low-memory submodules.\\
      & Lightweight Parameter-Efficient Tuning & \cite{cai2020tinytl, liu2022few, karimi2021compacter, wang2025loraedge, wang2025low,zhu2023pockengine, kwon2024tinytrain} & Minimal trainable layers for local optimization.\\
      & Prompt and Adapter Composition & \cite{wu2024apt, lester2021power, pfeiffer2021adapterfusion, ruckle2021adapterdrop,kwon2024tinytrain,lin2022device} & Modular adapters enable flexible task adaptation.\\
    \bottomrule
  \end{tabular}
  }
  \label{tab:compression}
\end{table*}

Deploying deep learning models on resource-constrained devices requires efficient compression techniques to reduce computational cost, memory footprint, and power consumption while maintaining optimal performance. This section explores key model compression methods (summarized in Table ~\ref{tab:compression}, including pruning, knowledge distillation, quantization, weight sharing, and Neural Architecture Search (NAS), with a focus on their role in optimizing the onboard learning process. These techniques enable efficient and lightweight models by eliminating redundancy, reducing precision, and optimizing architectures, ensuring deep learning models remain practical for real-world applications.

\subsection{Efficient Parameter Reduction: Pruning}

% \Figure[t!](topskip=0pt, botskip=0pt, midskip=0pt)[width=0.8\textwidth]{Fig_1_prune.pdf}
% { \textbf{General architecture of structured and unstructured pruning. (The dotted orange lines and circles denote neurons or connections to be pruned). In structured pruning, whole neurons or channels are removed to create smaller, hardware-efficient networks, while unstructured pruning discards individual connections to produce sparse models suited for low-resource computation.}\label{fig1}}

Pruning is a fundamental technique for onboard learning that reduces the number of parameters in a neural network, enabling deep learning models to operate within the memory, computational, and energy constraints of resource-limited devices~\cite{cheng2024survey, li2022optimizing, liu2022bringing}. By eliminating less important weights, pruning reduces model size while preserving accuracy, making it highly suitable for real-time applications. The pruning process assigns importance scores to parameters, removing those with the lowest values based on a predefined pruning rate. Selecting an optimal pruning rate is critical to maintaining accuracy while minimizing computational overhead.

\subsubsection{Unstructured vs. Structured Pruning.} Pruning can be broadly classified into {unstructured} and {structured} approaches~\cite{xia2022structured}. Unstructured pruning removes individual weights across the network, leading to sparse weight matrices that can improve efficiency when supported by dedicated accelerators. However, since standard deep learning frameworks and hardware do not efficiently support sparse computation, unstructured pruning often requires specialized hardware or software libraries for performance gains. In contrast, structured pruning removes entire groups of weights, such as neurons, channels, or filters, reducing model dimensions in a way that aligns well with conventional deep learning frameworks. Structured pruning is generally more suitable for onboard learning, as it produces a compact, hardware-friendly model compatible with edge devices~\cite{yu2020easiedge, ro2021autolr}.Structured pruning methods on ResNet architectures demonstrate 59.8-74.4\% computational reduction with 1.0-1.12\% accuracy loss while achieving 1.84-2.93$\times$ inference speedup on edge hardware including NVIDIA Jetson TX2, Snapdragon 855, and Google TPU~\cite{li2022optimizing, kong2023towards, zhang2025optimization}. In contrast, unstructured pruning achieves higher compression ratios (93.75-98\% sparsity)~\cite{yu2024edge,joardar2022realprune,yin2024workload}, with methods like u-Ticket demonstrating 98\% pruning rate on VGG-16 with minimal accuracy loss (91.0\% to 90.7\%) while improving hardware utilization and reducing operational latency by up to 76.9\%~\cite{yin2024workload}. When specialized support is available, unstructured methods like Edge-LLM demonstrate 4$\times$ memory reduction and 2.92$\times$ latency improvement with 1.29\% accuracy gain over baseline on LLaMA-7B~\cite{yu2024edge}. The overall architecture difference is illustrated in Figure ~\ref{fig1}.

\subsubsection{Pruning at Different Stages.} Pruning can be applied at various points in the model lifecycle: {before training} (pruning at initialization), {during training} (dynamic pruning), or {after training} (post-training pruning). Each stage offers distinct trade-offs in terms of computational efficiency and model accuracy.

\textit{\textbf{Pruning at Initialization.}} Pruning at initialization removes unnecessary parameters before training begins, reducing model complexity and enabling efficient training on resource-constrained devices~\cite{eccles2024dnnshifter, yu2020easiedge, hayou2020robust, jiang2022model}. A common approach is layerwise pruning, where entire layers or channels are removed based on their contribution to model performance~\cite{yu2020easiedge, ro2021autolr, zhu2023fedlp, li2022optlayer}. 

Several methods have been proposed to enhance pruning at initialization. SynFlow prevents layer collapse by selectively pruning while maintaining model trainability~\cite{tanaka2020pruning}. {Single-shot network pruning (SNIP)} assigns sensitivity scores to each weight and removes unimportant ones before training, ensuring critical connections are preserved~\cite{frantar2023sparsegpt, kohama2023single}. {CPSCA}, introduced by Liu et al.~\cite{liu2021channel}, employs spatial and channel attention to identify and prune unimportant channels while preserving inference accuracy. However, attention mechanisms in CPSCA introduce computational overhead, which may be prohibitive for highly constrained devices.

Another promising method is {channel clustering pruning}, proposed by Chang et al.~\cite{chang2023iterative}. This method groups channels based on feature similarity before pruning, optimizing the pruned network through knowledge transfer. Such approaches facilitate quicker performance recovery post-pruning and improve model efficiency.

Unstructured pruning at initialization focuses on weight-level sparsity, where individual parameters are removed based on significance~\cite{wang2022trainer, luo2020neural}. While computationally efficient, single-shot pruning may lead to accuracy degradation if not followed by iterative fine-tuning~\cite{xu2023efficient}. Together, these pruning techniques enable flexible and efficient model compression for onboard AI systems.

\textit{\textbf{Pruning During Training.}} Dynamic pruning modifies the model structure during training, selectively removing parameters based on real-time performance metrics. This technique can significantly reduce computational costs, with applications on edge devices demonstrating up to 94\% reduction in multiply-accumulate (MAC) operations, which directly lowers both time and space complexity of multi-head self-attention, resulting in 7.5-16$\times$ faster inference on edge devices with only 1-4\% accuracy degradation~\cite{jelvcicova2022delta}.

Moreover, {Adaptive pruning}, such as PruneFL~\cite{jiang2023computation}, dynamically resizes models during federated learning to balance communication and computation overhead while maintaining accuracy. Another approach named,{online pruning}, introduced by Guo et al.~\cite{guo2024online}, integrates real-time pruning into the training process, eliminating near-zero parameters to reduce computational load and improve inference efficiency. These techniques enable onboard AI to dynamically optimize model complexity without significant performance degradation, making them ideal for resource-limited environments.

\textit{\textbf{Post Training Pruning.}}In this compression approach, pruning is performed after model training, aiming to improve deployment efficiency without additional retraining. It selectively removes parameters with minimal contribution, such as weakly influential weights or channels, and uses a small calibration dataset to fine-tune activation consistency, which ensures minimal accuracy loss while avoiding full gradient-based optimization.
To tackle the information loss by structured pruning, a two-phase reconstruction process based on post-training pruning (PTP) was proposed in \cite{li2025improving}, which was able to reduce FLOPs by 1.73x on  ResNet-50 with only 3.55\% accuracy degradation compared to baseline using as little as 0.2\% of ImageNet data as a calibration dataset. This pruning approach first reduced information loss by sparsifying pruned channels and then optimized the reconstruction loss after pruning to obtain output signal from each layer. Similarly, layer-wise calibration approaches have demonstrated up to 65\% sparsity in ResNet50 at 8-bit precision with a ~1\% Top-1 accuracy drop, supporting deployment on commodity CPUs and embedded systems \cite{lazarevich2021post}. For language models, PTP has proven equally effective in resource-constrained hardware setup. 
In \cite{kwon2022fast}, structured PTP is demonstrated for transformers-based architectures, such as BERT and DistilBERT, which have achieved 2$\times$ FLOPs reduction and 1.56$\times$ faster inference while maintaining near original accuracy without retraining. Moreover, structured PTP methods designed for on-device LLMs have shown that hybrid-granularity pruning combined with mask tuning can retain over 91.2\% of baseline performance at a 20\% pruning ratio, while reducing memory usage by up to 80\% for the LLaMA-2-7B model \cite{xu2025lightweight}.

\Figure[t!](topskip=0pt, botskip=0pt, midskip=0pt)[width=0.9\textwidth]{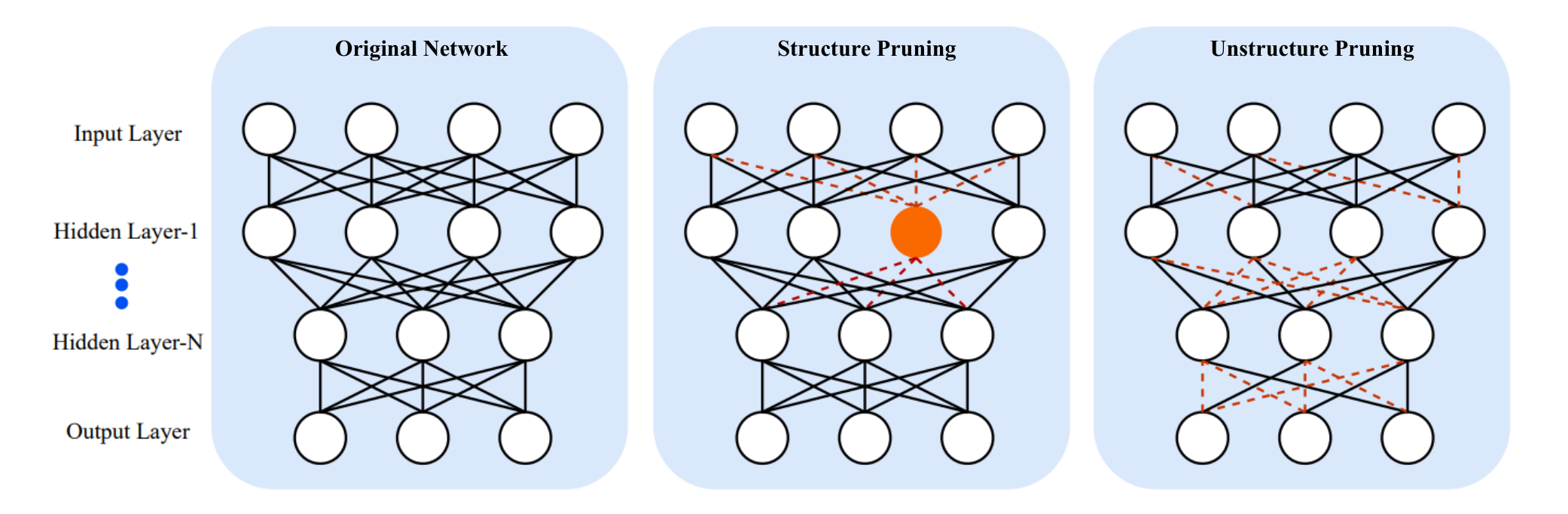}
{%
  \begin{minipage}[t]{0.8\textwidth} % <-- [t] makes the box top-aligned
    \raggedright
    \textbf{General architecture of structured and unstructured pruning. (The dotted orange lines and circles denote neurons or connections to be pruned). In structured pruning, whole neurons or channels are removed to create smaller, hardware-efficient networks, while unstructured pruning discards individual connections to produce sparse models suited for low-resource computation.}
  \end{minipage}%
  \label{fig1}%
}

\subsection{Precision Optimization: Quantization}

Quantization is a precision optimization technique that compresses neural network parameters and activations by mapping high-precision floating-point tensors to low-bit integer formats. This transformation reduces both memory footprint and computational overhead while maintaining inference accuracy, making it essential for efficient on-device learning and deployment in resource-constrained environments~\cite{hong2022daq,motetti2024joint,zhu2024towards}. The general quantization of a real-valued tensor \(v_\text{float}\) to an integer tensor \(v_\text{int}\) can be expressed as Eq. ~\ref{eq:quant_general} \& ~\ref{eq:quant_affine}.
\begin{equation}
v_\text{int} = \text{clip}\!\left(\text{round}\!\left(\frac{v_\text{float} - v_{\min}}{s}\right) + p,\, p,\, q\right)
\label{eq:quant_general}
\end{equation}
where \(v_{\min}\) and \(v_{\max}\) represent the tensor’s dynamic range, \([p,q]\) is the representable integer range (e.g., \([-128,127]\) for INT8), and \(s = (v_{\max} - v_{\min}) / (q - p)\) defines the quantization step size. The corresponding dequantization, which reconstructs the real-valued estimate, follows an affine mapping:
\begin{equation}
\hat{v}_\text{float} = s \cdot (v_\text{int} - z),
\label{eq:quant_affine}
\end{equation}
where \(z = \text{round}(p - v_{\min}/s)\) is the zero-point aligning the integer and real zero. Together, Equations~\ref{eq:quant_general}-\ref{eq:quant_affine} form the foundation of asymmetric affine quantization used in deployment frameworks such as TensorFlow Lite~\cite{jacob2018quantization} and ONNX~\cite{wu2020integer}.

For illustration (adapted from~\cite{jacob2018quantization}), consider a per-tensor INT8 quantization applied to a convolutional weight tensor with range \([-0.5, 0.8]\). With \([p,q] = [-128,127]\), the computed scale is \(s = (0.8 - (-0.5))/255 = 0.0051\), and the zero-point is \(z = 98\). A weight value \(v_\text{float} = 0.25\) thus maps to \(v_\text{int} = \text{round}(0.25/0.0051) + 98 = 147\) and reconstructs as \(\hat{v}_\text{float} = 0.0051 \times (147 - 98) = 0.25\). This concise affine formulation minimizes reconstruction error and supports energy-efficient integer arithmetic which is required for adaptive optimization in on-device AI systems~\cite{zhang2018efficient,fan2020training,wang2022edcompress}.

\subsubsection{Post-Training Quantization (PTQ).} PTQ applies quantization to a trained model without additional retraining, converting FP32 tensors to INT8 while maintaining accuracy~\cite{jeon2022mr,kim2024towards}. A single scaling parameter adjusts the quantization range per layer or channel, ensuring adaptability to different data distributions. Fine-tuning based on calibration samples further refines accuracy. However, PTQ may struggle to generalize across unseen data distributions, leading to potential precision loss. Online dynamic schemes mitigate this by recalibrating tensor ranges during inference but introduce computational overhead~\cite{rokh2023comprehensive}.

\subsubsection{Quantization-Aware Training (QAT).} Unlike PTQ, QAT incorporates quantization into the training process to enhance performance on quantized hardware. During forward propagation, weights and activations are quantized, simulating deployment conditions~\cite{rasch2023hardware,shen2021once}. Backpropagation updates FP32 weights, minimizing quantization errors while training. This dual-format approach improves model robustness but introduces hyperparameter tuning complexities~\cite{nagel2022overcoming}.

Compared with PTQ, QAT achieves a balanced trade-off between accuracy and computational efficiency. Reported results indicate that QAT attains between 4-12$\times$ compression and 2-10$\times$ improvement in inference throughput, while main- taining accuracy within 3\% of full-precision models~\cite{zhang2025survey}. The efficiency gain arises from the use of low-bit integer arithmetic, which reduces both storage demand and memory bandwidth during inference. However, performance behavior diverges sharply below 4-bit precisionand accuracy begins to degrade nonlinearly as quantization noise exceeds the corrective capacity of gradient updates, causing instability and slower convergence. In contrast, latency tends to scale almost linearly with bit-width reduction, showing that computational speed is governed mainly by the number of arithmetic operations rather than model depth. Hardware-adaptive schemes such as QuantEdge further expand this framework by dynamically adjusting quantization levels during inference. Tests on Jetson AGX Xavier, Asus Tinker Edge, and Raspberry Pi platforms demonstrate reductions of around 31.8\% in latency and 42\% in memory use, with less than 1\% loss in accuracy~\cite{mahmudov2025quantedge}. Overall, QAT serves as a co-optimization mechanism that aligns model granularity, inference latency, and hardware constraints for efficient deployment in resource-limited systems.

\subsubsection{Hybrid Run-Time Quantization (HRQ).} Merging post-training Quantization calibration with runtime adaptability, hybrid runtime quantization dynamically adjusts quantization parameters during inference to improve trade-offs in performance \cite{zheng2025fedhq}.
One of the approaches of HRQ is KV-cache quantization, where the "key" and "value" tensors are compressed on-the-fly during token generation, which has emerged as a predominant enabler for deploying small and efficient LLMs on edge and mobile devices. For example, SubKV showed per-channel key and per-token value quantization with a Dynamic Window Quantization mechanism can reduce KV-cache memory by over 75\%, achieving 2-bit precision with minimal 0.31 perplexity degradation on sub-billion models for LLMs, such as MobileLLM-600M and TinyLLaMA-1.1B \cite{zeng2025subkv}. Similarly, Oaken et al., \cite{kim2025oaken} integrate online-offline hybrid KV-cache quantization within accelerator hardware, achieving a 1.58$\times$ throughput improvement and 70\% reduction in bitwidth while maintaining less than 1\% accuracy loss. For edge computing with FL, FedHQ is introduced by combining PTQ  and QAT to balance performance across heterogeneous edge devices, achieving 2.47$\times$ training speedup and up to 11.15\% accuracy improvement compared with single-strategy baselines, while adding only 1.7 minutes of extra overhead \cite{zheng2025fedhq}.

\subsubsection{Mixed Precision Quantization (MPQ).} MPQ mitigates accuracy loss in ultra-low precision models by assigning different bit widths to different network components~\cite{koryakovskiy2023one}. While traditional INT8 quantization preserves accuracy in critical layers, MPQ improves compression and efficiency for onboard deployment. The challenge lies in optimally selecting bit widths, prompting techniques like OneShot Mixed-Precision Search ~\cite{koryakovskiy2023one}, which improved model correlation score  (measured using Kendall’s Tau approach) by 41\% compared with sampled child models and standalone fine-tuned model demonstrated in \cite{guo2020single} on the ImageNet dataset. Building on these developments, bit-width allocation in mixed precision quantization is guided by systematic sensitivity analysis rather than manual adjustment. The primary objective is to preserve model accuracy while reducing computational and memory costs by assigning precision levels according to each layer’s relative importance within the network. Layers with a higher loss curvature or greater gradient variance are considered more sensitive to quantization and therefore assigned higher precision to maintain numerical stability \cite{dong2019hawq,yao2021hawq}. However, less sensitive or computationally intensive layers are represented with lower precision, such as INT4, which reduces inference latency by 23\% compared to uniform INT8 quantization while maintaining 76.73\% top-1 accuracy on ImageNet. Moreover, recent approaches have formulated this process as a constrained optimization problem that balances accuracy, latency, and energy efficiency. For example, Hardware-Aware Quantization (HAQ) \cite{wang2019haq, zhao2025hardware} employs reinforcement learning to determine optimal precision configurations under device specific constraints, while Bayesian Bit-Width Quantization \cite{van2020bayesian} applies probabilistic inference to model the trade-off between representational fidelity and hardware cost.
Besides, Hessian Aware Quantization (HAWQ) introduced second-order MPQ, achieving an 8$\times$ compression on ResNet-20 with 1\% higher accuracy on CIFAR-10 and 68\% top-1 accuracy on ImageNet with a 1MB model~\cite{dong2019hawq}. MPQ offers a balance between accuracy and efficiency, making it highly suitable for onboard learning.

By integrating PTQ, QAT, and MPQ, onboard systems optimize model efficiency while maintaining performance, enabling real-time inference under stringent resource constraints.

\subsection{Knowledge Transfer for Compact Models: Knowledge Distillation}

% \Figure[t!](topskip=0pt, botskip=0pt, midskip=0pt)[width=0.85\textwidth]{Fig_2_KD.pdf}
% { \textbf{General architecture of knowledge distillation the student learns from both teacher soft labels and ground-truth data, achieving compact models with minimal loss in accuracy.}\label{KD}}

\Figure[t!](topskip=0pt, botskip=0pt, midskip=0pt)[width=0.9\textwidth]{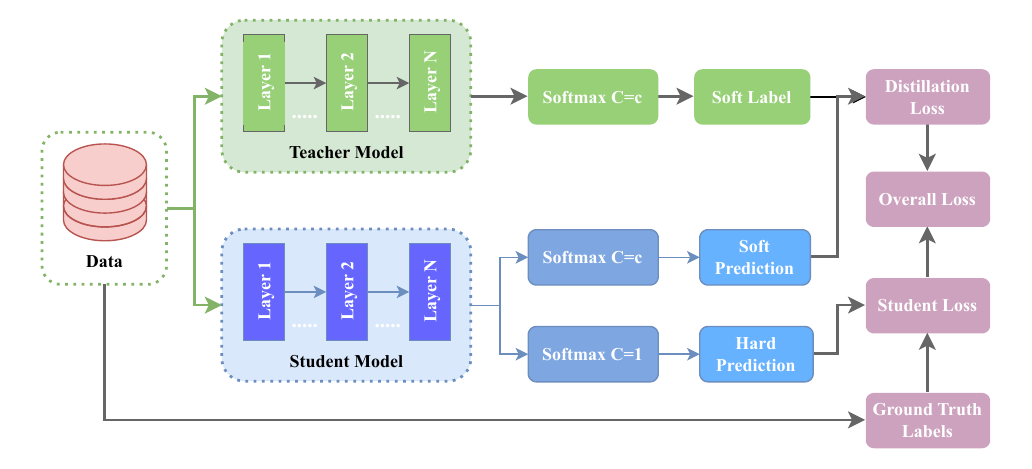}
{%
  \begin{minipage}[t]{0.8\textwidth} % <-- [t] makes the box top-aligned
    \raggedright
    \textbf{General architecture of knowledge distillation, where the student model learns from both the teacher’s soft labels and the ground-truth hard labels, yielding a compact model with minimal degradation in accuracy.}
  \end{minipage}%
  \label{KD}%
}

Deploying deep learning models on resource-constrained edge devices presents significant challenges due to limited computational power, memory, and energy efficiency~\cite{huang2022knowledge, yang2024survey}. Knowledge distillation (KD) addresses these limitations by transferring knowledge from a high-capacity teacher model to a lightweight student model, enabling efficient onboard deployment~\cite{gou2021knowledge, wang2021knowledge, zhu2021student}. In knowledge distillation (shown in Figure \ref{KD}), the student network is trained under two complementary forms of supervision: hard labels that convey the true class identities, and soft labels derived from the teacher’s temperature-scaled output distribution. The hard labels ensure alignment with ground-truth data, while the soft labels transfer the teacher’s nuanced knowledge of class relationships, enabling the student to approximate the teacher’s predictive behavior with greater generalization efficiency ~\cite{park2021learning}. This technique allows smaller models to retain high accuracy while reducing memory footprint and inference latency.

During training, the student model minimizes a loss function that integrates both hard labels (ground truth) and soft labels from the teacher. Soft labels encode inter-class relationships, providing richer supervision and enabling the student model to generalize better. However, since well-trained teacher models tend to produce sparse probability distributions with high confidence in the correct class, a temperature parameter $C$ is introduced in the softmax function to soften the distribution and enhance knowledge transfer. Specifically, given an input $\mathbf{X}$ with corresponding ground truth label $\mathbf{y}$, and teacher logits ${z_i}_{i=1}^{K}$ over $K$ classes, the softened soft label distribution is computed as:
$
q_i = \frac{\exp(z_i / C)}{\sum_{j=1}^{K} \exp(z_j / C)},
$
where $z_i$ represents the teacher’s logit for class $i$ and $C$ is the temperature parameter. As $C$ increases, $\mathbf{q}$ approaches a uniform distribution, enhancing inter-class information. The student model is optimized using a combined loss function $L_{\text{KD}} = \frac{1}{R} \sum_{\mathbf{X}, \mathbf{y}} \left( (1 - \gamma) L_{\text{Hard}}(\mathbf{X}, \mathbf{y}) + \gamma L_{\text{Soft}}(\mathbf{X}, \mathbf{q}) \right)$ where $\gamma$ controls the trade-off between the two loss terms, and $R$ is the total number of input samples. This loss function balances hard label loss $L_{\text{Hard}}$ and soft label loss $L_{\text{Soft}}$. The individual loss functions for general knowledge distillation are defined as $L_{\text{Hard}}(\mathbf{X}, \mathbf{y}) = -\sum_{i=1}^{K} y_i \log p_i(\mathbf{X})$ and $L_{\text{Soft}}(\mathbf{X}, \mathbf{q}) = -\sum_{i=1}^{K} q_i \log p_i(\mathbf{X})$, where $p_i(\mathbf{X})$ represents the student model’s predicted probability for class $i$. Hard labels use $C = 1$, whereas soft labels utilize the teacher’s logits.

The choice of distillation temperature $C$ has been shown to substantially influence student performance in on-device learning scenarios. For instance, in MobileNet-V3 distilled for edge deployment, accuracy improves from $94.3\%$ (without KD) to $97.9\%$ at an intermediate temperature of $C=2.0$, whereas very low ($C < 1$) or high ($C > 3$) values degrade performance (e.g., $94.7\%$ at $C=5.0$) \cite{ji2024edge}. Similar observations were shown in \cite{li2023curriculum} demonstrate on ImageNet that their curriculum temperature scheduling improves top-1 accuracy from $70.26\%$ (vanilla KD with fixed $C$) to $70.83\%$ (+0.57\%) by gradually adapting $C$ during training. Likewise, Chi et al.~\cite{chi2023normkd} demonstrated that normalizing logits and using adaptive temperature yields consistent gains of $0.5$-$1.2\%$ on CIFAR-100 for compact models such as ResNet-20 and MobileNet. Collectively, these results indicate that moderate or adaptive choices of $C$ are most effective for lightweight architectures in edge computing, as they preserve inter-class similarity while maintaining confidence in the target class.

Recent studies demonstrated memory and computational onboard AI deployment through KD~\cite{jang2020knowledge, fang2020edgeke, mishra2023designing, blakeney2020parallel}. For instance, Sepahvand et al.~\cite{sepahvand2022teacher} introduced a tensor decomposition-based KD approach, achieving a 265.67$\times$ compression rate for ResNet-18 with minimal accuracy loss.  In \cite{crowley2018moonshine}, a method named Moonshine is developed with a similar student-teacher architecture to reduce resource and memory utilization, which resulted in a 40\% smaller model with 5.02\% less error than the baseline after compression. Xiao et al. \cite{xiao2024knowledge} introduced DQN-KD, applying knowledge distilation with reinforcement learning to minimize memory utilization, where it achieved 50.4\% fewer FLOPs (flops full form floating point operations per second) than baseline with 47.5\% parameter reduction.

To reduce the communication overhead during onboard deployment, Itahara et al. \cite{itahara2021distillation} demonstrated semi-supervised federated learning with knowledge distillation that deducted 99\% communication cost while maintaining similar accuracy compared to the benchmark via onboard local models’ outputs exchange between heterogeneous devices and optimizes local models. This proposed approach adapted KD for onboard deployment by transferring global knowledge from a teacher model to client models using soft labels, allowing clients to learn generalized patterns despite non-IID data. 
However, recent studies reveal performance degradation concerns under severe non-IID conditions through selective self-distillation methods, which achieve only 53.37\% accuracy under extreme heterogeneity compared to 73.38\% under moderate non-IID settings \cite{he2022learning}. Similarly, in \cite{jin2022personalized}, lightweight deployment with FKD in non-IID settings observed 6.36\% performance degradation in CIFAR-100 when client number increased from 20 to 100, but outperformed non-IID FL baselines, such as FedProx \cite{li2020federated}, which catastrophically drops to 13.32\%.

Moreover, Qu et al. \cite{qu2020quantization} demonstrated an adaptive quantized federated knowledge distillation approach for edge devices to address the bottleneck of communication cost and achieved a target accuracy of 83.35\% with 30\% labelled data in CIFAR-10 and IMDb non-IID data. Luo et al. \cite{luo2022keepedge} proposed KeedEdge with a deep neural network (DNN) with 5.86\% improvement in the student model through knowledge distillation with 14\% model reduction, aiming to lower complexity and latency for UAV positioning.

To address the challenges of heterogeneous computation, Qi et al. \cite{qi2022fedbkd} introduced bidirectional knowledge distillation to facilitate efficient and scalable onboard deployment for modulation classification in IoT-edge systems in federated learning framework. The proposed knowledge distillation approach of multi-teacher knowledge distillation, where the global network was regarded as a student network that unifies the heterogeneous knowledge from multiple teacher networks, enabled lightweight model updates by transferring essential insights between local and global models, minimizing communication costs and reducing memory and computational demands on devices.

Overall, Knowledge distillation provides a principled framework for training compact models capable of retaining the predictive performance of larger architectures while reducing memory, computational demands, and inference latency. By transferring soft labels from a teacher model, it encodes rich inter-class relationships, enabling efficient deployment in resource-constrained environments. This feature is essential for real-time applications, such as autonomous vehicles, where rapid, localized decision-making is crucial.
Furthermore, distillation allows flexible deployment across diverse devices by optimizing student models for specific capabilities, enhancing performance in heterogeneous environments. It also supports continual learning by enabling incremental updates without full retraining, reducing computational overhead. In summary, knowledge distillation ensures efficient, adaptive, and high-performing onboard models that are adaptable to the constraints of edge and embedded systems.

\begin{table*}[h!]
\centering
\caption{Comparison of major NAS strategies on edge/onboard devices.}
\label{tab:nas_hw_comparison}
\renewcommand{\arraystretch}{1.7}
\begin{tabular}{@{}lcccccc@{}}
\toprule
\textbf{NAS Method} & \textbf{Parameters (M)} & \textbf{Memory Efficiency} & \textbf{Latency (ms)} & \textbf{Target Hardware} \\
\midrule
DARTS~\cite{liu2018darts} & 3.3 & High overhead & -- & GPU only \\
Resource-Constrained NAS~\cite{lyu2021resource} & 1.1 & reduced via binary-path selection & 28.2, $\sim$35\% faster & Jetson Nano\\
ProxylessNAS~\cite{cai2018proxylessnas} & 5.7 & Moderate & 78 & Mobile GPU/CPU \\
OFA~\cite{cai2020once} & 2.6 & Low & 130 (RasPi4), 1057 (FPGA) & RasPi4, FPGA, mobile \\
On-NAS~\cite{kim2023device} & -- & 90\% less vs. DARTS & -- & Jetson TX2 \\
OnceNAS~\cite{zhang2024oncenas} & 1.97 & Low & 100 (RasPi4), 1046 (FPGA) & RasPi4, FPGA, CPU \\
TinyNAS~\cite{lin2020mcunet} & 1.2 & $<$320 KB SRAM & 95-129 for different model & ARM Cortex-M7 MCU \\
HW-NAS (ML-based)~\cite{juneja2025accelerating} & -- & Weight sharing, efficient search & 11.87 & Jetson, Coral TPU \\
\bottomrule
\end{tabular}
\end{table*}

\subsection{Automated Architecture Optimization: Neural Architecture Search}
The deployment of deep learning models on edge devices presents challenges due to diverse hardware architectures, real-time processing demands, and stringent resource constraints. Unlike traditional deep learning models optimized for high-performance GPUs or cloud environments, onboard AI must balance accuracy, latency, and energy efficiency~\cite{ye2024deep}. Neural Architecture Search (NAS) has emerged as a powerful model compression approach, automating the design of efficient deep learning architectures tailored to memory-constrained embedded devices~\cite{chitty2022neural}. By dynamically optimizing models for specific hardware, NAS accelerates the design process and enhances adaptability, making it a key enabler of next-generation edge AI.

Traditional deep learning models often require manual adaptation to fit the stringent constraints of embedded platforms, a process that is both labor-intensive and suboptimal for latency critical applications. To overcome this, resource-aware NAS has emerged as a systematic alternative. For instance, Lyu et al.~\cite{lyu2021resource} proposed a multi-objective NAS that directly optimizes on devices such as Jetson Nano and Raspberry Pi, achieving Pareto-optimal architectures with up to 35\% latency reduction and 25\% memory savings compared to hand-tuned baselines \cite{zhou2024hgnas}.

However, early gradient-based NAS methods such as DARTS \cite{liu2018darts} remain impractical for edge deployment, as their differentiable supernet training introduces substantial memory overheads and cannot be executed within the limits of devices like Raspberry Pi or Jetson Nano. More efficient approaches emerged to address this. ProxylessNAS \cite{cai2018proxylessnas} reduced memory consumption by binarized path sampling, enabling deployment on mobile GPUs cutting earch cost by 200$\times$ and obtained 3.1\% higher Top-1 accuracy (comapared to MobileNetV2) and 6$\times$ fewer parameter on CIFAR-10 dataset, while Once-for-All (OFA) \cite{cai2020once} achieved 75\% parameter reduction from 2.6M to 10M and inference latency as low as 130 ms on Raspberry Pi 4, which fastens the overall deployment and learning onboard. At the extreme low-power end, TinyNAS with TinyEngine~\cite{lin2020mcunet, lin2023tiny} that employed two-stage NAS with topology level memory scheduling, delivering 4.1$\times$ lower SRAM use than ProxylessNAS \cite{cai2018proxylessnas} and up to 1.7$\times$ aster inference than TF-Lite Micro \cite{zaidi2022unlocking} as well as executed in less than 320 KB SRAM on Cortex-M micro-controllers, enabling resource-efficient onboard learning through practical real-time inference.

More recent work explicitly integrates hardware constraints into NAS objectives. On-NAS~\cite{kim2023device} demonstrated 90\% lower peak memory on embedded GPUs by expectation-based operation selection with pretraining meta cell into single condensed via two-fold-meta learning, while OnceNAS~\cite{zhang2024oncenas} unified latency and parameter count in its search process, demonstrating 1.97M parameters and 100 ms latency on Raspberry Pi 4 with superior accuracy compared to EdgeNAS and ProxylessNAS. Similarly, ML-based HW-NAS~\cite{juneja2025accelerating} introduced a predictive modeling framework to accelerate architecture search, achieving 11.87ms latency and 51.16 mJ energy consumption for specific candidate setup on Jetson Nano with 50 billion MACs. This framework reduces the architecture search time from days to hours, enabling efficient on-device deployment (Jetson Nano and Edge Coral TPU)  by predicting latency and energy consumption without requiring exhaustive hardware evaluations.

Overall, Table~\ref{tab:nas_hw_comparison}  illustrates a clear progression with early NAS methods like DARTS are GPU-centric and resource-prohibitive, while hardware-aware NAS frameworks (OFA, On-NAS, OnceNAS, HW-NAS) achieve efficient, deployable models across Raspberry Pi, Jetson, FPGA, and MCU platforms. This trend highlights that the future of onboard NAS lies in multi-objective, hardware-aware, and energy-efficient strategies.

In conclusion, NAS provides a systematic framework for compressing and optimizing deep learning models for onboard AI, enabling automated adaptation to onboard hardware constraints. By leveraging hardware-aware and few-shot NAS methods, onboard systems can achieve efficient, high-performance compressed model design while minimizing memory usage and computational overhead.

\subsection{Adaptive Compression: Adapter-Based Fine-Tuning}
\label{subsec:adapter_based_compression}

General model compression techniques primarily aim to reduce model size and computational cost prior to deployment. While effective for static inference scenarios, such approaches offer limited flexibility once models are deployed, particularly under dynamic, non-stationary, or user-specific data distributions common in edge environments. Adapter based fine-tuning addresses this limitation by introducing lightweight, trainable modules into otherwise frozen pretrained networks, enabling post-deployment adaptation with minimal overhead. In this paradigm, the vast majority of parameters remain fixed, while a small subset—typically below 1--2\% of the total model—is updated during fine-tuning. This design enables gradient-based local learning under tight memory and energy constraints, preserves data privacy by avoiding raw data transmission, reduces communication overhead in decentralized or federated settings, and supports rapid task- or user-specific personalization directly on-device.

\textbf{Parameter-Efficient Adaptation Methods.}
Adapter based fine-tuning has developed into a broad class of parameter-efficient learning methods that exploit low-rank structure, sparsity, and quantization. Low-Rank Adaptation (LoRA) represents a core technique by factorizing weight updates into two low-rank matrices, achieving up to a 10,000-fold reduction in trainable parameters, approximately 0.01\% of the full model, together with around $3\times$ GPU memory savings. Importantly, this reduction is achieved while maintaining performance comparable to full fine-tuning and without introducing inference-time latency~\cite{hu2022lora}. For convolutional architectures that dominate edge vision workloads, LoRAE extends this principle through the introduction of LoRA-extractor and LoRA-mapper modules that better preserve spatial structure. As a result, trainable parameters are reduced to roughly 4\% of full fine-tuning requirements~\cite{wang2025loraedge, wang2025low}. Experiments on YOLOv8x report parameter reductions of 86.1\%, 98.6\%, and 94.1\% for image classification, object detection, and segmentation, respectively, with rank-8 object detection retaining competitive accuracy despite a 98.6\% reduction in trainable parameters.

Quantized LoRA (QLoRA) further improves efficiency by combining low-rank updates with 4-bit NormalFloat quantization of frozen weights. This enables fine-tuning of 65B-parameter models on a single 48\,GB GPU while achieving more than $16\times$ memory efficiency and performance reaching 99.3\% of ChatGPT-level baselines~\cite{dettmers2023qlora}. Related methods, including LQ-LoRA and L4Q, integrate sub-8-bit quantization with low-rank optimization to preserve accuracy under aggressive precision constraints, making them suitable for energy- and memory-limited hardware~\cite{bondarenko2024low, guo2023lq, jeon2024l4q}. Beyond low-rank updates, scaling-based approaches such as IA$^3$ introduce per-layer multiplicative scaling factors for attention and feed-forward modules, resulting in less than 0.1\% parameter overhead~\cite{liu2022few}. Compacter further reduces overhead by employing shared Kronecker-factorized adapter matrices, achieving approximately $100\times$ compression while maintaining strong generalization across tasks~\cite{karimi2021compacter}. Prompt-based techniques, including Prompt-Tuning and Prefix-Tuning, adapt frozen models by inserting small trainable vectors or tokens into the input embedding space, thereby steering model behavior with minimal parameter cost~\cite{lester2021power, wu2024apt}. Runtime-oriented strategies such as AdapterFusion and AdapterDrop further reduce computational cost by reusing or selectively skipping adapters during inference, with AdapterDrop reporting approximately 39\% faster multi-task inference while maintaining near-parity latency for single-task settings~\cite{pfeiffer2021adapterfusion, ruckle2021adapterdrop}.

\textbf{Activation Memory Optimization.}
Although parameter-efficient updates substantially reduce model size, activation memory remains a major bottleneck for on-device training, often consuming three to five times more memory than model parameters during backpropagation~\cite{cai2020tinytl}. Tiny-Transfer-Learning (TinyTL) directly targets this issue by freezing feature extractors and updating only bias terms and lightweight residual components, thereby eliminating the need to cache large activation tensors. This approach achieves 6.5-12.9$\times$ reductions in training memory without measurable accuracy loss, and in several cases yields substantial improvements over last-layer fine-tuning~\cite{cai2020tinytl}. Across nine image classification benchmarks, TinyTL reduces training memory from 391\,MB to 37\,MB on ProxylessNAS-Mobile and from 850\,MB to 66\,MB on Inception-V3, while delivering a 34.1\% accuracy improvement relative to naive fine-tuning~\cite{cai2020tinytl}. Extending this line of work, TinyTrain enables adaptive compression on microcontroller-class platforms through task-adaptive sparse training and dynamic layer and channel selection. Operating within a 1\,MB memory budget, TinyTrain demonstrates $9.5\times$ faster training and $3.5\times$ lower energy consumption on Raspberry Pi Zero 2~\cite{kwon2024tinytrain}. For even more constrained platforms with sub-256\,KB SRAM, system and algorithm co-design approaches combining quantization-aware scaling and sparse updates enable deployment on devices such as STM32F746, achieving 20--21$\times$ memory savings and 23--25$\times$ speedups while matching cloud-trained accuracy~\cite{lin2022device}.

\textbf{System-Level Integration.}
System-level optimizations are critical for translating adaptive compression methods into practical on-device learning pipelines, particularly under irregular computation patterns induced by sparsity and selective parameter updates. PockEngine exemplifies this direction through compile-time backward-graph pruning guided by layer-wise importance profiling. This approach achieves up to $15\times$ speedup over TensorFlow on Raspberry Pi and $5.6\times$ memory savings on Jetson AGX Orin~\cite{zhu2023pockengine}. As a result, large models such as LLaMA2-7B can be fine-tuned on-device at 550 tokens per second, representing a $7.9\times$ improvement over PyTorch-based implementations. Overall, the combination of parameter-efficient adaptation methods, activation-memory optimization techniques, and system-level compilation strategies establishes adaptive compression as a scalable and practical foundation for privacy-preserving and energy-efficient edge intelligence across hardware platforms ranging from data-center GPUs to deeply embedded microcontrollers, without dependence on continuous cloud connectivity.

\textbf{Overview.}
Across the compression and optimization studies reviewed in this section, several quantitative trends emerge that are directly relevant for onboard deployment. Structured pruning methods achieve substantial efficiency gains, with reported reductions in multiply-accumulate (MAC) operations of up to 94\% on edge hardware while remaining within 1-4\% of baseline accuracy, and sparsity levels around 65\% in ResNet-50 at 8-bit precision with roughly 1\% Top-1 accuracy loss. Post-training quantization further reduces computation and memory, with individual studies showing up to 80\% memory savings and latency reductions of about 23\% on embedded platforms while sustaining accuracy above 76.7\%. These examples indicate that, when pruning ratios and bit widths stay within hardware-supported limits, compression can cut computation and storage very aggressively with only modest accuracy degradation.

Knowledge distillation (KD) complements these techniques by recovering performance lost due to pruning and quantization. In the compressed settings discussed in this section, KD improves accuracy from 94.3\% to 97.9\% with less than 10\% additional compute, and similar studies report that distilled students retain accuracy close to the original models under domain shift and reduced training budgets. Hardware-aware neural architecture search (NAS) targets the same trade-offs at design time: NAS-derived models achieve around 35\% inference speed-up on Jetson Nano and up to 90\% reduction in search cost compared with conventional DARTS-style baselines, while maintaining competitive accuracy. Parameter-efficient fine-tuning approaches, such as TinyTL and LoRAEdge, then enable onboard adaptation by updating less than 5\% of the parameters or a small set of low-rank adapter modules, so that inference latency and energy remain close to those of the frozen backbone.

Overall, these reported results suggest a practical interaction pattern. Pruning and quantization largely determine whether a model is feasible on a given device in terms of memory, computation, and latency. KD is used to restore or improve accuracy within that compressed resource envelope. NAS incorporates these constraints at model-design time by producing hardware-matched architectures. Parameter-efficient fine-tuning finally provides a mechanism for continual or adaptive learning without retraining full models. This view links concrete numerical results from the literature to a simple decision guideline in which the choice of optimization technique is driven by whether energy and latency, accuracy, or adaptability is the dominant constraint on the target onboard platform.

\Figure[t!](topskip=0pt, botskip=0pt, midskip=0pt)[width=0.9\textwidth]{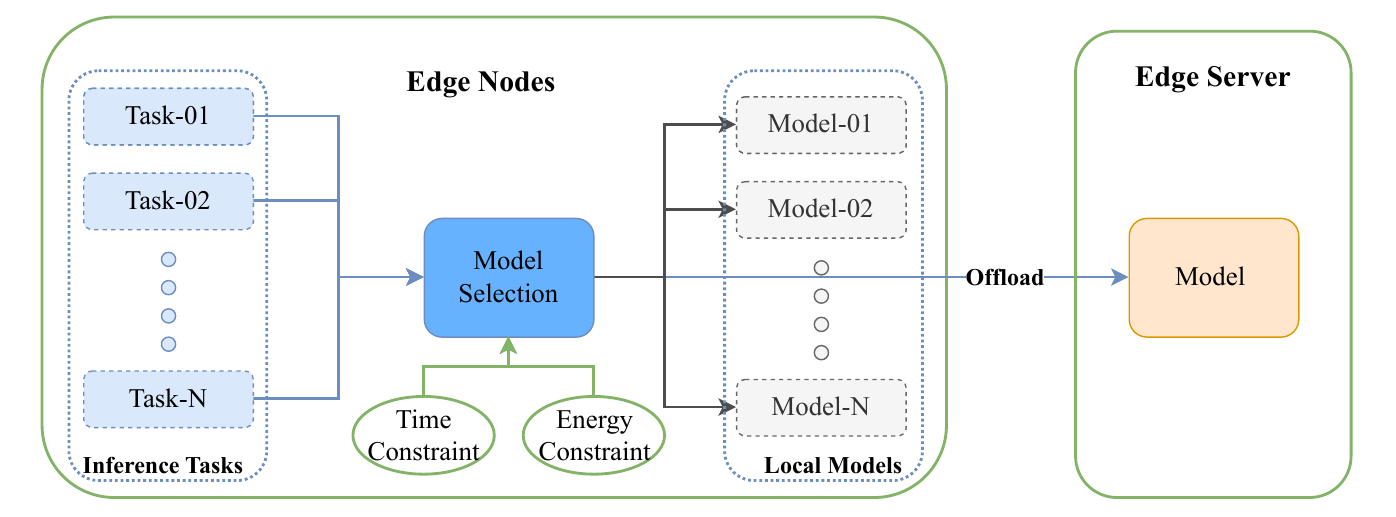}
{%
  \begin{minipage}[t]{0.8\textwidth} % 
    \raggedright
    \textbf{General architecture of edge-based computational offloading., where edge nodes manage multiple inference tasks by selecting appropriate local models based on latency and energy constraints, while complex computations are dynamically offloaded to edge servers to enhance efficiency and responsiveness.}
  \end{minipage}%
  \label{inference_offloading}%
}

\section{Efficient Inference} \label{sec:efficient_inference}

To achieve efficient inference on resource-constrained edge devices, the combination of optimization strategies plays a vital role. In onboard setup with low processing capability, computation offloading enables efficient inference through computationally intensive tasks delegated to external resources, reducing the burden onboard. While offloading is limited by network constraints or privacy concerns, model partitioning offers a viable alternative by distributing computations between the device and external systems to enhance inference for execution. Moreover, in onboard execution, where responsiveness and minimal energy consumption are required, early exit strategies further optimize inference by allowing models to terminate processing once a confident prediction is reached. This section highlights these techniques in detail by discussing their benefits and trade-offs in achieving efficient inference for optimized onboard processing.

% \begin{figure*}[t]
%   \centering
%   \includegraphics[width=0.8\textwidth]{Fig_3_computionaloffloading.pdf}
%   \caption{General architecture of edge-based computational offloading.}
%   \label{inference_offloading}
%   \vspace{-10pt}
% \end{figure*}

\subsection{Distributed Computation for Onboard AI: Computation Offloading}

% \Figure[t!](topskip=0pt, botskip=0pt, midskip=0pt)[width=0.85\textwidth]{Fig_3_computionaloffloading.pdf}
% { \textbf{General architecture of edge-based computational offloading. edge nodes manage multiple inference tasks by selecting appropriate local models based on latency and energy constraints, while complex computations are dynamically offloaded to edge servers to enhance efficiency and responsiveness.}\label{inference_offloading}}

Computation offloading (shown in Figure ~\ref{inference_offloading}) enables resource-constrained devices to execute deep learning tasks efficiently by delegating computationally intensive operations to nearby edge servers. This approach reduces energy consumption and improves inference speed, which is critical for real-time applications such as autonomous navigation and surveillance~\cite{duan2021computation, jeong2018computation, huda2023deep, kim2018dynamic, xu2020trust, shakarami2021autonomous}. While network latency and security risks remain concerns, offloading enhances scalability and optimizes resource utilization across distributed systems~\cite{abbas2021computational, dey2019offloaded}. 

To improve inference efficiency, DNNs can leverage offloading strategies that selectively assign tasks between local devices and edge servers. Zhou et al.~\cite{zhou2022accelerating} proposed a fused layer-based model parallelism approach that dynamically partitions DNN layers in multi-access edge computing (MEC) environments using Particle Swarm Optimization, achieving a 12.75$\times$ inference speedup. Similarly, Teerapittayanon et al.~\cite{teerapittayanon2017distributed} introduced a distributed DNN framework where edge devices execute partial inference, aggregating intermediate outputs to minimize communication overhead. To further optimize offloading, Nikoloska et al.~\cite{nikoloska2020data} developed a data selection scheme that transmits only uncertain samples, while Fresa et al.~\cite{fresa2021offloading} employed LP-relaxation-based dynamic programming for optimal task scheduling.

Several techniques improve offloading efficiency by reducing computational load and bandwidth requirements. Deep compressive offloading, introduced by Yao et al.~\cite{yao2020deep}, applies compressive sensing to reduce transmitted data size, achieving a 2-4$\times$ reduction in latency with minimal accuracy loss. Gao et al.~\cite{gao2020rethinking} proposed entropy-based pruning to selectively remove redundant features, further improving efficiency. Additionally, runtime management strategies dynamically adjust computational priorities, bypassing less critical computations to minimize latency~\cite{li2019edge}. 

When a single device lacks sufficient resources, collaborative edge computing enables multiple devices to function as a virtual edge server, distributing inference tasks dynamically~\cite{xue2020edgeld}. This cooperative framework balances computational load, reduces latency, and enhances real-time responsiveness by leveraging shared computational resources across edge devices.

Recent advances in onboard inference demonstrate that efficient computation is also achievable through specialized neural processing units (NPUs). Modern mobile system-on-chips (SoCs) now integrate NPUs capable of high-throughput INT8 operations (e.g., 73 TOPS on Qualcomm Hexagon), enabling fast on-device inference for large models. Offloading the prefill \cite{patel2024splitwise} stages of LLMs to NPUs, while handling floating-point operations and activation outliers, achieving 30.7$\times$ energy savings and 22.4$\times$ faster prefill compared to CPU/GPU-only inference ~\cite{xu2025fast}. Moreover, mllm-NPU LLM is proposed for prefilling acceleration with on-chip NPU offloading, introducing chunk-wise CPU-NPU coscheduling and dynamic outlier inference \cite{xu2024wip}. Such hybrid execution highlights how computation offloading can extend beyond cloud-edge cooperation to the hardware level, optimizing performance within a single device through heterogeneous processing units.

\subsection{Adaptive Execution Strategies: Model Partitioning}

\Figure[t!](topskip=0pt, botskip=0pt, midskip=0pt)[width=0.8\textwidth]{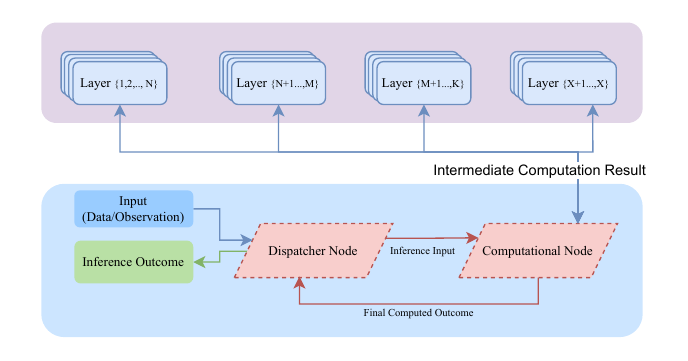}
{%
  \begin{minipage}[t]{0.8\textwidth} % <-- [t] makes the box top-aligned
    \raggedright
    \textbf{Model partitioning for deep learning inference at the edge, where computations are distributed across dispatcher and computational nodes, allowing early layers to execute locally while deeper layers run remotely. This adaptive partitioning reduces latency and energy use while maintaining real-time performance.}\\[2pt]
  \end{minipage}%
  \label{partitioning}%
}

Model partitioning optimizes onboard inference by distributing computation across multiple devices, ensuring low latency, energy efficiency, and real-time responsiveness~\cite{feltin2023dnn, chow2023krisp}. Instead of fully offloading computations, partitioning allows selective execution of neural network layers, retaining early-stage processing onboard while delegating high-complexity computations to external resources~\cite{liang2023dnn}. This technique balances computational demand while minimizing data transfer overhead.

Layer-wise partitioning assigns inference computations at specific layer boundaries, enabling lightweight feature extraction locally while deeper layers refine representations on an edge server. Operator-based partitioning further decomposes models into computational units, dynamically distributing inference tasks for efficient execution. To enhance adaptability, adaptive partitioning adjusts inference assignments in real-time based on network conditions, processing power, and latency constraints~\cite{chow2023krisp}. Quantitative analysis across heterogeneous inference deployments reveals consistent inverse relationships between bandwidth availability and optimal partition depth. Kang et al.~\cite{kang2017bottlenet} profiled eight DNN architectures across mobile-cloud configurations, demonstrating 3.1$\times$ speedup and 59.5\% energy reduction relative to cloud-only execution. As network upload latency degrades from WiFi (95 ms) through LTE (180 ms) to 3G (870 ms), optimal partition points migrate from intermediate network layers toward complete cloud offloading, reflecting bandwidth-constrained optimization.

Memory-communication trade-offs exhibit nonlinear scaling with partition granularity in distributed IoT inference deployments. For instance, Zhao et al.~\cite{zhang2018deepsplit} partitioned ResNet-50 and YOLOv2 inference across Raspberry Pi 3 clusters via Fused Tile Partitioning (FTP), achieving 30-50 ms inference latency with accuracy within 1\% of centralized baselines. Memory footprint reduction scales from 58\% to 68\% as grid dimension increases from 3$\times$3 to 5$\times$5, while inference communication overhead grows proportionally from 13.8 MB to 18.9 MB. This contrasts sharply with layer-wise partitioning exhibiting linear communication growth around 14.0-19.6 MB across 1-6 devices. Moreover, distributed work stealing reduces inference data movement by 52\%, enabling 1.7-3.5$\times$ speedup across 2-6 devices through peer-to-peer task transfer that eliminates centralized coordination bottlenecks.

Dynamic inference partitioning strategies adapted to network variability demonstrate superlinear performance improvements relative to static configurations. Hu et al.~\cite{hu2019dynamic} formulated partition selection as min-cut optimization for DAG-structured networks, achieving 6.45$\times$ and 8.08$\times$ inference latency reductions versus edge-only and cloud-only execution respectively, with throughput gains of 8.31$\times$ and 14.01$\times$. Mohammed et al.~\cite{mohammed2020distributed} extended multi-partition distribution across fog networks, demonstrating 2.6-4.2$\times$ inference speedup through context-aware workload assignment that accounts for heterogeneous device capabilities. Hierarchical partitioning considering core-level heterogeneity yields further inference optimization, such as, HiDP ~\cite{taufique2025hidp} achieved 38\% latency reduction, 46\% energy savings, and 56\% throughput improvement across EfficientNet, InceptionNet, ResNet, and VGG inference on commercial edge platforms including Jetson Orin NX, Jetson Nano, and Raspberry Pi variants.

Despite its advantages, model partitioning introduces (shown in Figure~\ref{partitioning}) trade-offs between latency, accuracy, and energy consumption. While offloading reduces onboard computation, unstable network conditions may introduce latency overhead. Additionally, frequent inter-device communication can impact efficiency, necessitating optimized scheduling strategies. By integrating intelligent partitioning with adaptive exit mechanisms, onboard AI achieves a balance between computational efficiency and inference performance, making it a critical approach for real-time onboard learning.

\subsection{Dynamic Stopping for Faster Inference: Early Exit Strategies}

Early exit mechanisms optimize onboard inference by allowing deep learning models to terminate computation once a confident prediction is reached, reducing latency and energy consumption~\cite{laskaridis2021adaptive, teerapittayanon2016branchynet}. This technique is particularly beneficial for resource-constrained devices, where real-time responsiveness is critical. By skipping unnecessary layers, early exits enhance efficiency without sacrificing accuracy, making them essential for fast and adaptable onboard/edge inference.

Early exit strategies enable models to process input data through initial layers and stop inference if confidence exceeds a predefined threshold~\cite{passalis2020efficient}. If confidence remains low, additional layers may be executed locally or offloaded to an edge or cloud server for further refinement~\cite{li2023predictive, dong2022resource}. This flexibility minimizes computational overhead while maintaining predictive reliability.

When integrated with computation offloading and model partitioning, early exits further optimize resource allocation. Initial layers process data onboard, and if a confident prediction is achieved, offloading is bypassed, reducing latency and bandwidth usage~\cite{li2023predictive, pacheco2021towards, jo2023locoexnet}. Otherwise, deeper layers are computed remotely to refine predictions. 

Furthermore, adaptive approaches enhance early exits by dynamically selecting the optimal exit layer based on real-time conditions. Reinforcement learning (RL) improves efficiency by learning optimal exit points based on past inference outcomes~\cite{she2024dynamic, wolczyk2021zero}. An RL agent determines when to stop computation, balancing speed and accuracy to match dynamic workloads. Federated learning (FL) further refines early exit models by enabling collaborative training across multiple edge devices without sharing raw data, preserving privacy while improving robustness~\cite{douch2024early, ilhan2023scalefl, zhong2022flee}.

% By integrating early exits with offloading, RL, and FL, onboard inference becomes more efficient, reducing computation, minimizing energy consumption, and improving adaptability to dynamic workloads while maintaining data privacy.

While both early exit and model compression have individually advanced the efficiency of deep inference onboard, most prior work has evaluated these approaches in isolation, thereby overlooking their synergistic potential. As discussed in Section~\ref{sec:model_compression}, compression techniques reduce the computational and memory footprint of each layer, where early exit strategies, by contrast, adaptively terminate inference once a sufficiently confident prediction is reached, reducing the average depth of computation \cite{behera2025exploring}. When jointly applied, these mechanisms operate orthogonal dimensions compression that lowers the cost of computation per layer, whereas early exit decreases the number of layers executed, resulting in a compounded effect that exceeds the gains of either method alone \cite{korol2025iot}. For example, in \cite{shi2019improving}, the integration of two-step pruning with edge device cooperative inference demonstrated a reduction in total latency by up to 4.8$\times$ and accelerates computation by 6$\times$ compared to uncompressed baselines was shown in \cite{laskaridis2020spinn}. This combined deployment was also able to deduct transmission workload by 25.6$\times$ and validated that progressive early exit scheduling across device and cloud enhances throughput by 2$\times$ and lowers server cost by 6.8$\times$ under dynamic network conditions. Moreover, Predictive Exit was introduced in \cite{li2023predictive} by coupling quantized (INT8) inference with dynamic voltage-frequency scaling, achieving 96.2\% computation and 72.9\% energy reduction, and outperforming conventional early-exit schemes by an additional 37.6\% in energy savings. Moreover, Dong et al. \cite{dong2022multi} demonstrated that co-optimizing multi-exit strategies with compression and partitioning in the MAMO framework accelerates edge inference by up to 13.7$\times$ with minimal accuracy loss. Kutukcu et al. \cite{kutukcu2024slexnet} showed that integrating width-adaptive compression and depth-adaptive early exits achieves superior trade-offs between under latency power dynamic execution conditions. In general, these findings confirm that compression statically optimizes the model structure while early exit dynamically adapts the runtime depth, forming a cohesive optimization paradigm that amplifies the efficiency of inference across the dimensions of latency, energy, and bandwidth.

\textbf{Overview.}
Efficient onboard inference depends on how computation placement, execution depth, and model complexity are coordinated under device and network constraints. Computation offloading provides the largest latency and energy benefits under stable connectivity, with reported speedups exceeding one order of magnitude and energy reductions approaching an order of magnitude on heterogeneous edge platforms, whereas degraded bandwidth consistently shifts optimal execution toward earlier partition points or fully local inference~\cite{kang2017bottlenet,zhou2022accelerating,hu2019dynamic}. Model partitioning repeatedly appears as the mechanism that mediates this transition by trading increased communication for reduced onboard computation, achieving stable latency improvements in the range of $3\times$--$8\times$ and energy savings of up to $59.5\%$ across mobile and embedded devices~\cite{kang2017bottlenet,zhang2018deepsplit,taufique2025hidp}. Early exit strategies further reduce average inference depth, enabling substantial computation and energy reductions (reported up to $96.2\%$ and $72.9\%$, respectively) by avoiding unnecessary deep execution for easy inputs~\cite{li2023predictive,laskaridis2021adaptive}. Reported results indicate that compression and early exit are complementary, jointly lowering per-layer cost and executed depth, which amplifies the effectiveness of partitioned and offloaded inference, leading to multi-fold latency reductions ($4.8\times$-$13.7\times$) and significant reductions in data transfer volume~\cite{shi2019improving,laskaridis2020spinn,dong2022multi}. Overall, these quantitative trends indicate that deployable onboard inference arises not from any single optimization technique, but from the coordinated use of offloading, partitioning, early exit, and compression to balance accuracy, latency, and energy under changing deployment conditions.

\section{Decentralized Learning} \label{sec:decentralized_learning}

Decentralized learning is crucial for onboard AI, enabling edge devices to train and update models collaboratively without relying on centralized servers. By distributing learning across multiple devices, it enhances privacy, reduces latency, and improves adaptability in dynamic environments. However, deploying decentralized learning on resource-constrained devices introduces several challenges, including data heterogeneity, model adaptation over time, and computational efficiency under dynamic resource constraints. This section systematically examines decentralized learning in onboard AI, categorizing existing methods based on the key challenges they address.

\subsection{Privacy-Preserving Model Updates: Federated Learning}

Federated Learning (FL) is a foundational strategy in decentralized learning, enabling edge devices to collaboratively train shared models without exposing private local data~\cite{lim2020federated,ye2020edgefed,su2021secure}. By exchanging only model updates, FL reduces privacy leakage and lowers communication costs, which is vital for deploying onboard intelligence across resource-constrained edge environments~\cite{wang2023coopfl,kim2021autofl}. However, FL faces core limitations in such settings, including non-IID data distributions, communication bottlenecks, and the challenge of continual model adaptation under tight hardware, bandwidth, and energy constraints~\cite{tao2023byzantine,liao2023accelerating,wang2019adaptive}. These limitations have motivated a growing body of work on decentralized and lightweight FL paradigms that better align with the requirements of onboard learning and optimization.

\textbf{Incremental and locally adaptive model updates.}
One of the key research directions focuses on enabling incremental learning and local adaptability within decentralized FL frameworks. Ito et al.~\cite{ito2021device} proposed an on-device FL approach based on Online Sequential Extreme Learning Machines (OS-ELM) combined with autoencoders for unsupervised anomaly detection. Unlike backpropagation-based FL, their method updates models sequentially using streaming local data and performs cooperative model fusion through one-shot merging of intermediate results. Evaluations show ROC-AUC scores of 0.99 on MNIST and 0.96 on HAR, with consistent performance on UAH-DriveSet, demonstrating that non-iterative, lightweight neural models can achieve competitive accuracy for continual onboard learning under decentralized operation~\cite{ito2021device,chen2022decentralized,liu2025accelerating}. These results challenge the notion that deep or computationally intensive architectures are necessary for effective model adaptation on edge.

\textbf{Communication-efficient model exchange.}
Communication efficiency remains a dominant bottleneck in decentralized FL, particularly for large models and bandwidth-limited edge networks. Liu et al.~\cite{liu2025accelerating} introduced CEDFL, a decentralized FL framework with probabilistic peer-to-peer communication, allowing devices to dynamically select neighbors based on bandwidth availability and model divergence. Their results show an 11\% improvement in test accuracy and a 55\% reduction in training completion time compared to deterministic decentralized baselines. In parallel, Chen et al.~\cite{chen2022decentralized} proposed exchanging intermediate representations instead of full gradients or parameters, achieving up to 81\% reduction in communication cost while preserving convergence behavior. Complementarily, Asheralieva et al.~\cite{asheralieva2025dynamic} developed a dynamic distributed model compression strategy, enabling devices to adjust compression ratios based on network dynamics and computation availability. Their results indicate improved convergence and reduced straggler effects, reinforcing the role of communication-aware design for real-time onboard optimization.

\textbf{System-level decentralization and deployability.}
Beyond algorithmic design, practical onboard learning requires frameworks that can operate without centralized control. Zhang et al.~\cite{zhang2025enabling} proposed EdgeFL, a fully decentralized FL architecture that eliminates the need for a parameter server and enables direct peer-to-peer aggregation. Experiments on MNIST and CIFAR-10 demonstrate faster model convergence and reduced latency compared to centralized and semi-centralized baselines, all while requiring minimal developer integration overhead. The EdgeFL framework supports asynchronous participation and flexible model sharing, addressing practical deployment challenges in dynamic edge environments~\cite{zhang2025enabling,lim2020federated,wang2019adaptive}. This positions decentralized FL as a viable engineering solution for production-grade onboard AI.

\textbf{Scalability through multi-model and asynchronous collaboration.}
To address heterogeneity in resource availability and data distribution, MMDFL~\cite{yan2025mmdfl} introduces a decentralized FL paradigm based on asynchronous "traveler" models that circulate among nodes, aggregating insights without requiring synchronized rounds. Experimental results show a 10.7\% accuracy gain and 30\% reduction in communication rounds compared to conventional decentralized FL, especially under non-IID data settings. This strategy mitigates model divergence and supports long-term adaptation in AIoT and embedded systems, where consistent connectivity and hardware uniformity are rare~\cite{yan2025mmdfl,liu2023blockchain,kim2021autofl}.

\textbf{Security and personalization for robust deployment.}
Robustness and personalization remain essential for decentralized onboard FL. Tao et al.~\cite{tao2023byzantine} developed a Byzantine-resilient training algorithm combining distributed SGD with gradient compression, maintaining convergence even with malicious clients in the network. Meanwhile, personalized FL frameworks like CoopFL~\cite{wang2023coopfl} and AutoFL~\cite{kim2021autofl} adapt to device-specific data through local fine-tuning and architectural adjustments. These methods enable collaborative training while ensuring relevance to local environments, which is crucial for user-facing applications such as mobile health, autonomous vehicles, and smart sensors.

In summary, federated learning has evolved from a centralized privacy-preserving paradigm into a decentralized, adaptive, and deployment-aware framework for onboard learning and optimization. Advances in on-device incremental learning~\cite{ito2021device}, communication-efficient and compressed model exchange~\cite{liu2025accelerating,asheralieva2025dynamic}, system-level decentralization~\cite{zhang2025enabling}, and multi-model learning~\cite{yan2025mmdfl} collectively address the intertwined challenges of data heterogeneity, resource constraints, and continual adaptation. These developments establish decentralized FL as a practical foundation for real-time, embedded, and resource-aware AI systems.

\subsection{Distributed Training: Split Learning}

Split Learning (SL) is a distributed training paradigm in which a deep neural network is partitioned between an edge device and a coordinating node, allowing each to execute separate segments of the model collaboratively without exchanging raw data \cite{hu2025review}. In this architecture, the device processes early network layers using its local sensor data, then transmits intermediate activations to the server for completion of the forward and backward passes. Only gradient updates for the cut layer are returned, preserving data privacy and substantially reducing device-side computation and communication load \cite{lin2024efficient}. SL represents a pivotal advancement in enabling decentralized onboard intelligence beyond the limitations of conventional FL. While FL decentralizes gradient aggregation, it still demands full local model training, imposing substantial computational and energy burdens on constrained devices. In contrast, SL structurally partitions the neural network between client and coordinating nodes, exchanging only intermediate activations rather than raw data or complete gradients. 

The efficiency of this structural paradigm has been quantitatively validated across diverse real-world deployments. For instance, Li et al. \cite{li2024introducing} demonstrated that implementing federated split learning on real smart-meter hardware with 192 KB of SRAM reduced the memory footprint by 95.5\%, shortened training time by 94.8\%, and decreased communication volume by 50\%, while maintaining forecasting accuracy comparable to server-trained models. In industrial Internet of Things environments, Li et al. \cite{li2024fsledge} reported around 38\% lower communication delay and 27\% reduction in energy consumption using an energy-aware federated split learning framework (FSLEdge) without notable accuracy degradation. 

Despite its advantages, split learning suffers from slower performance than FL, as only one client interacts with the shared model segment at a time, leaving other clients idle and increasing training overhead as the number of edge clients grows. To address this, Splitfed Learning was proposed with refined architecture incorporating with PixelDP and differential privacy, making it suitbale for fast training and where global model based on a continually updating dataset over time was required \cite{thapa2022splitfed}.In distributed collaborative approach and testing in several benchmarks, it achieved around 99.2\% accuracy on MNIST and improved 1.7\% comparing with FL in HAM10000 dataset, establishing it as a more efficient decentralized alternative for onboard learning.

\subsection{Continual Adaptation in Decentralized Systems: Continual Learning}

Continual Learning (CL) enables AI models to incrementally acquire new knowledge while preserving past information, making it critical for real-time onboard learning. Unlike traditional FL, which assumes fixed training tasks, CL allows models to evolve dynamically. However, CL faces two key challenges: catastrophic forgetting and computational efficiency in resource-limited environments.

\textbf{Mitigating Catastrophic Forgetting.} Standard CL methods suffer from catastrophic forgetting, where previously learned knowledge deteriorates as new tasks are introduced. To address this, LightCL~\cite{wang2024comprehensive, buzzega2020dark} freezes the lower and middle layers of a pre-trained model while updating only higher layers, reducing memory consumption while preserving past knowledge. Additionally, task-aware dynamic masking (TDM) selectively retains critical parameters, balancing adaptation and retention.

\textbf{Decentralized Continual Learning.} Continual learning in decentralized settings introduces additional challenges due to device heterogeneity and intermittent connectivity. Decoupled replay mechanisms, such as Chameleon~\cite{aggarwal2023chameleon}, maintain separate short-term and long-term memory buffers, optimizing data retention while minimizing resource usage. Further improvements are achieved through hardware-accelerated optimizations, as demonstrated in Chameleon-Accel~\cite{pellegrini2020latent, ravaglia2021tinyml}, which achieves up to 2.1$\times$ speedup over SLDA and 3.5$\times$ over Latent Replay, making CL feasible for onboard deployment.

\textbf{Cross-Edge Continual Learning.}  
Traditional FL frameworks assume static task contexts and fixed device associations. However, mobile edge devices often traverse multiple federated domains over time, requiring CL mechanisms that can generalize across learning environments. Cross-FCL~\cite{zhang2022cross} addresses this by decomposing neural parameters into shared and task-specific subspaces. Only shared parameters are communicated during FL rounds, preserving task-specific knowledge locally while avoiding interference during aggregation. On CIFAR-100 and Office-Home benchmarks, Cross-FCL improves average accuracy by up to 17.2\% over baseline FCL methods under non-IID conditions, while maintaining low memory overhead~\cite{zhang2022cross}. The framework's ability to isolate task-relevant features and prevent degradation during cross-edge transfers makes it particularly suited for decentralized continual learning in real-world multi-domain edge deployments.

\textbf{Hardware-Aware Continual Learning.} Efficient continual learning on resource-constrained devices requires approaches that jointly optimize accuracy, memory footprint, latency, and energy consumption. Ravaglia et al.~\cite{ravaglia2021tinyml} present a quantized latent replay (QLR) framework for on-device continual learning on TinyML platforms. By compressing latent replay buffers using 8-bit quantization, the proposed method achieves a fourfold reduction in memory usage with only a 0.26\% accuracy loss on the Split MNIST benchmark when using 3000 latent replay samples. On the VEGA 22\,nm prototype platform, the system operates 65$\times$ faster and is 37$\times$ more energy efficient than a low-power STM32L4 microcontroller, enabling continual learning within a total memory budget below 64\,MB and reudcing 4.5$\times$ needed for rehersaling. Moreover, evaluating in non-centralized environments, Li et al.~\cite{li2025unleashing} demonstrated an accuracy improvement of 10.36\% over FedAvg on CIFAR-10 Class-Incremental Learning tasks, while incurring only a 1.97\% increase in communication rounds through modular federated continual learning approaches. These findings confirm that with targeted algorithmic and hardware-level optimizations, continual learning can be both scalable and sustainable for real-time, onboard intelligence in decentralized edge systems.

While CL enhances adaptability, it still operates under the assumption of relatively stable resource availability. However, in highly dynamic environments, onboard AI must continuously adjust its learning strategy based on fluctuating constraints, requiring adaptive learning techniques.

\subsection{Dynamic Adaptation Under Resource Constraints: Adaptive Learning}

Adaptive learning extends continual learning by dynamically adjusting model behavior based on real-time changes in computational resources, task complexity, and environmental variations. In edge AI scenarios, where resources fluctuate, learning strategies must be both efficient and responsive.

\textit{Real-Time Model Adjustment.} IMPALA~\cite{espeholt2018impala} and SEED RL~\cite{espeholt2019seed} employ asynchronous updates and parallel experience collection, enabling models to efficiently learn from decentralized environments while reducing synchronization bottlenecks. Importance-weighted corrections ensure stability by mitigating inconsistencies in decentralized updates~\cite{engstrom2020implementation}.

\textit{Resource-Aware Learning Strategies.} Jin et al.~\cite{jin2021learning} proposed a resource-aware optimization framework that formulates a non-linear mixed-integer program to dynamically allocate computational resources for learning tasks. Their method reduces communication overhead and optimizes onboard switching between edge devices, ensuring efficient adaptation in dynamic environments.
\textit{Low-Rank and Parameter-Efficient Adaptation.} Parameter-efficient adaptive methods exploit low-rank structures to decouple shared and agent-specific knowledge while minimizing redundancy in decentralized systems. Zhang et al.~\cite{zhang2025low} proposed the Low-Rank Agent-Specific Adaptation (LoRASA) framework, which embeds low-rank adapters into a shared policy backbone to enable scalable coordination among agents in multi-agent reinforcementt learning. On SMAC and MAMuJoCo benchmarks, LoRASA reduced parameter counts by 45-65\%, while maintaining task performance within 1\% of non-parameter-sharing baselines. These findings demonstrate that low-rank adaptation provides an effective balance between scalability, efficiency, and coordination for edge-deployed multi-agent systems.

\textit{Scalability and Fault Tolerance in Decentralized Systems.} By prioritizing localized learning and reducing reliance on central servers, decentralized approaches enhance scalability, improve latency efficiency, and strengthen fault tolerance. These properties make decentralized learning particularly suitable for autonomous edge systems, IoT networks, and federated AI applications. Through dynamic resource allocation and distributed model updates, adaptive learning ensures robust and efficient AI decision-making in continuously evolving real-world settings. 

\textit{Hardware-Aware Optimization for Adaptive Onboard Learning}
Hardware-aware optimization extends the principles of adaptive learning to the circuit level, ensuring that deep models can execute efficiently on embedded and reconfigurable devices. This class of techniques co-designs neural architectures and hardware implementations -leveraging quantization, distributed arithmetic, and matrix compression to deliver real-time inference with minimal latency and power overhead. Such designs transform algorithmic adaptability into deployable intelligence, allowing deep networks to operate autonomously within the limited computational envelopes of edge hardware.

Chen et al. \cite{chen2025comet} introduced COMET, a CNN co-optimization framework employing Offset-Binary Coding (OBC) for both inputs and weights to reduce arithmetic complexity in FPGA implementations. Using a modified LeNet-5 model on an Xilinx FPGA, COMET replaced multiply-accumulate operations with shift-accumulate logic and optimized lookup-table designs, achieving substantial reductions in logic and memory utilization with negligible accuracy loss. In the recurrent domain, Khan et al. \cite{khan2024digit} proposed a digit-serial distributed-arithmetic RNN architecture fabricated in 65 nm CMOS, achieving 74.54\% smaller core area, 68.66\% lower power, and 2.61$\times$ higher throughput than earlier OBC-based RNN implementations, confirming the scalability of arithmetic-level optimization for ASIC systems. 

Expanding these techniques to temporal learning models, Alhartomi et al. \cite{alhartomi2023low} applied distributed arithmetic to LSTM networks using circulant and block-circulant matrix vector multiplications for compression. Implemented on FPGA hardware, their design achieved 98.71\% DSP reduction, 33.59\% fewer lookup tables, 13.43\% fewer flip-flops, and 29.76\% power reduction compared with state-of-the-art LSTM accelerators. Similarly, Yalamarthy et al. \cite{yalamarthy2019low} presented a pipelined, distributed-arithmetic LSTM on a Xilinx Zynq-7000 FPGA, incorporating circulant matrix compression to reduce area by 74.5\% and power by 68.6\%, achieving 3.89$\times$ higher hardware efficiency than multiplier-based designs.

Thus, these studies demonstrate that co-optimization between learning algorithms and hardware accelerators, spanning CNN, RNN, and LSTM models, enables real-time, energy-efficient onboard deployment. When coupled with decentralized adaptive learning strategies, such hardware-aware designs bridge the gap between algorithmic intelligence and physical execution, paving the way for fully autonomous, resource-aware edge systems.

\textbf{Overview.}
Federated, split and continual learning on resource-limited devices inherit the design constraints established by compression and inference optimization. A compressed 4-bit model reduces uplink communication volume by 70-80\% \cite{li2020federated} but increases susceptibility to drift, leading to 2-4\% accuracy degradation after multiple rounds without compensation from distillation or replay. Replay-based continual learners \cite{ro2021autolr} regain stability at the cost of 10 \% extra energy per epoch, while parameter-efficient methods such as LoRAEdge \cite{wang2025loraedge} confine updates to adapter layers, cutting per-round computation by 60 \%. These figures show that model efficiency, adaptability, and stability form a coupled triad: greater compression improves communication efficiency but weakens continual retention; adapter-based or distilled updates restore performance but extend training time. Quantitatively balancing this triad is essential for sustained onboard learning under intermittent connectivity and limited power budgets.

\section{Security and Privacy} \label{sec:security_privacy}

The deployment of deep learning models on edge devices introduces significant security and privacy challenges~\cite{chen2019deep}. Unlike centralized cloud-based AI, onboard learning operates in resource-constrained environments, making it susceptible to adversarial attacks, data breaches, and inefficiencies in privacy-preserving mechanisms~\cite{gill2025edge}. These vulnerabilities enhance risks to model integrity, confidentiality, and robustness, necessitating dedicated security measures. This section systematically examines key security threats and mitigation strategies, focusing on three fundamental aspects: privacy protection in decentralized learning, secure model execution, and trustworthy AI for onboard learning (summarized in Table ~\ref{tab:security_privacy}.

\begin{table*}[t]
\caption{Security and Privacy Threats in Onboard AI}
\renewcommand{\arraystretch}{2}
\centering
\resizebox{\textwidth}{!}{
\begin{tabular}{p{2.5cm} p{2cm} p{3.5 cm} p{5.2cm} p{3.5cm} p{4.3cm}}
\toprule
\textbf{Threat} & \textbf{Attack Type} & \textbf{Attack Methods} & \textbf{Countermeasures} & \textbf{Overhead} & \textbf{Edge-Specific Pitfalls} \\
\midrule

Data Leakage from Model Updates & Privacy &
Gradient inversion, attribute inference~\cite{wang2021membership,zhao2020idlg, bai2024membership} &
Differential Privacy, 
gradient pruning, encrypted checkpoints~\cite{abadi2016deep,chen2022decentralized,bonawitz2017practical,el2022differential} &
3\% to 27\% accuracy loss and increase in training time~\cite{abadi2016deep} &
Heterogeneous clients limit uniform DP, constrained compute for noise sampling. \\

Membership Inference & Privacy &
Presence inference on training sets~\cite{wang2021membership,bai2024membership} &
Local DP-SGD, randomized response, secure aggregation~\cite{abadi2016deep,bonawitz2017practical} &
Low protocol overhead with FL, high tolerance to failing devices~\cite{bonawitz2017practical} &
Non-uniform privacy budgets across devices. \\

Adversarial Inversion & Privacy / Confidentiality &
Reconstruction from training or sensitive features~\cite{tan2025defending,zhao2020idlg} &
Noise-based regularization, backdoor defense, DP masking for split inference~\cite{tan2025defending, li2023gan} &
Additional computational cost for training indicator tasks &
Backdoor attacks via model update poisoning. \\

Side-Channel Attacks & Hardware / Execution &
Cache Prime-Probe~\cite{lou2021survey}, DRAM timing attack~\cite{kwong2020rambleed} &
Data perturbation and DP~\cite{xiao2019edge,ansari2020security}, noise injection~\cite{wang2021enabling,van2019tale}, layer-wise computations and encrypted model execution~\cite{wang2021enabling, moini2021power} &
2.1\% power overhead with path delay of 0.62 ns~\cite{bhade2024lightweight} &
Hard to synchronize cache randomization on low-power SoCs. \\

Memory/Timing Side-Channels & Hardware / Physical &
Memory-access and timing leaks, weight revealing~\cite{hua2018reverse} &
Hide memory access patterns using ORAM and avoiding dynamic zero pruning optimizations~\cite{hua2018reverse} &
Constant-access designs add significant overhead &
Overhead leaves accelerators still vulnerable. \\

Adversarial Perturbation & Integrity &
Gradient-based targeted~\cite{ozbulak2020perturbation}, white/black-box variant and ensemble AutoAttack &
Layered or attention-guided distillation training + bidirectional metric learning~\cite{wang2021agkd,mehta2024layered} &
Multi-step adversarial training computationally heavy as it requires many backprop passes &
May cause gradient masking, falsely inflating robustness on resource-constrained hardware. \\

Update Poisoning (Local) & Robustness / Integrity &
Toxic sample points via gradient ascent direction~\cite{zhu2023research} &
Differentially private and generative federated learning~\cite{hitaj2017deep,zhang2020poisongan} &
Low communication, moderate server computation &
Filtering costs on low-resource devices, sensitivity to non-IID data. \\

Inference Data Exposure & Privacy &
Exposure of inputs/intermediates during split or partitioned inference, model stealing, membership inference~\cite{tramerslalom,zhang2024no,hu2022membership} &
Hybrid TEE frameworks (Slalom~\cite{tramerslalom}, ShadowNet~\cite{sun2023shadownet}), encrypted offloaded tensors (ENIGMA~\cite{li2022enigma}, TEE-Shielded DNN Partition~\cite{zhang2024no}), Lyapunov-based methods~\cite{cheng2025privacy} &
40\% transmission delay but higher privacy loss with increasing storage~\cite{cheng2025privacy} &
Untrusted accelerator offloading exposes intermediate features. \\

Model Extraction Attack & Security / Confidentiality &
Weight stealing via DeepSteal, chained attack flow~\cite{rakin2022deepsteal} &
Encrypted model blobs, weight-bit masking, substitute model training with mean-cluster weight penalty~\cite{gong2021model,rakin2022deepsteal} &
Around 24\% accuracy drop during attack analysis~\cite{rakin2022deepsteal} &
Most significant bit (MSB) leakage causes high latency, impractical for resource-constrained hardware. \\

Compression \& Quantization Privacy Leakage & Privacy &
Attack on features or class membership from pruned/quantized parameters~\cite{liu2025quantization,yuan2022membership} &
DP-based quantization, randomized rounding, sparsity with noise, entropy-based pruning, secure aggregation~\cite{liu2025quantization,yuan2022membership,becking2020ecq,boselli2025explainable} \& security-aware pruning (X-Pruner)~\cite{yu2023x} &
Encoding/decoding and cryptographic overhead increase per-round latency &
Randomization must accompany compression to avoid ranking leakage. \\

Transparency / XAI Exposure & Interpretability / Privacy &
Leakage via saliency maps, gradients, explanation artifacts~\cite{liang2021explaining,zhang2024survey} &
Federated/DP explainability (FED-XAI), DP explanation sharing under privacy budgets, layer-wise relevance propagation \& SHAP~\cite{patel2022model,bechini2023application,cantini2024xai,boselli2025explainable} &
22-50\% encrypted traffic classification with XAI~\cite{senevirathna2025enhancing} &
Compute limits for XAI on micro edges, explanation leakage risks. \\
\bottomrule
\end{tabular}
}
\label{tab:security_privacy}
\end{table*}

\subsection{Privacy Protection in Decentralized Learning}

In distributed and federated edge learning, data privacy remains a fundamental concern. Because model updates often contain information about the underlying data, attackers can infer sensitive details through gradient inversion or membership inference techniques~\cite{wang2021membership,zhao2020idlg,bai2024membership}. For example, the iDLG framework~\cite{zhao2020idlg} can reconstruct ground-truth labels if gradients from individual samples are exposed, illustrating how raw model updates can reveal private content.

To reduce this leakage, Differential Privacy (DP) has become a widely adopted strategy~\cite{abadi2016deep,bonawitz2017practical,chen2022decentralized,el2022differential}. DP introduces controlled random noise into gradients or model parameters, thereby obscuring individual contributions. However, this approach comes at a cost: reported accuracy reductions range from 3\% on small datasets such as MNIST to 27\% on complex image tasks like CIFAR-10~\cite{abadi2016deep}. Training times also increase because of the additional noise sampling operations. These effects are amplified in heterogeneous edge networks, where computational power and noise generation capacity vary widely between devices.

To complement DP, secure aggregation protocols~\cite{bonawitz2017practical} enable model updates to be combined without revealing any single participant’s data. They are lightweight and tolerant of device failures, but the lack of a uniform privacy budget across clients can still expose weaker nodes. Similarly, adversarial inversion methods~\cite{tan2025defending,li2023gan} can reconstruct data from intermediate features during split learning. Defenses based on noise regularization and DP masking reduce this risk, though they introduce noticeable computational overhead for each training iteration.

Privacy threats also emerge from compression and quantization processes~\cite{liu2025quantization,yuan2022membership}. When parameters are pruned or quantized, subtle correlations can still leak class or feature information. Secure compression schemes such as DP quantization, randomized rounding, and entropy-based pruning~\cite{becking2020ecq,boselli2025explainable,yu2023x} have been proposed to mitigate this issue. These methods improve confidentiality but add communication latency and increase per-round computation time. In addition, transparency mechanisms such as Explainable AI (XAI) pose a new category of privacy risk. Visualization tools including saliency maps or SHAP values can unintentionally expose information about the training distribution~\cite{liang2021explaining,zhang2024survey}. To balance interpretability and privacy, Federated Explainability (FED-XAI) frameworks~\cite{patel2022model,cantini2024xai,boselli2025explainable} combine DP with constrained explanation sharing. Despite these efforts, traffic analysis studies still show that 22-50\% of encrypted network flows can be correctly classified~\cite{senevirathna2025enhancing}, demonstrating that full privacy under interpretability remains an unresolved challenge.

\subsection{Secure Model Execution and Adversarial Defenses}

While privacy-preserving mechanisms protect user data during decentralized learning, the security of the execution environment itself is equally critical. Onboard and edge AI systems must safeguard both model integrity and execution confidentiality against adversarial manipulation and side-channel exploitation. Hardware-based security architectures such as Trusted Execution Environments (TEEs) have emerged as foundational solutions for this purpose. Platforms like Intel SGX, ARM TrustZone, and AMD SEV provide isolated environments for sensitive computations, ensuring that encryption keys, model parameters, and intermediate activations remain inaccessible to untrusted processes~\cite{ohrimenko2016oblivious,mofrad2018comparison,manzoor2024survey,demigha2021hardware}.
TEE-assisted frameworks such as TEE-ML~\cite{hesamifard2018privacy} extend these principles to deep learning, enabling encrypted model inference and training with minimal impact on performance. These environments act as hardware enclaves that preserve confidentiality even in shared or adversarial hardware setups, establishing a hardware root of trust for onboard learning.

Despite these benefits, hardware accelerators and peripheral devices introduce vulnerabilities that adversaries can exploit through side-channel attacks (SCAs) and timing analysis. Edge hardware often leaks sensitive information through power consumption traces, cache timing variations, and memory access sequences~\cite{wang2021enabling,lou2021survey,kwong2020rambleed}.
Furthermore, power analysis-based inversion can reconstruct neural network weights by monitoring FPGA power patterns~\cite{moini2021power}, while cache-based timing attacks can infer layer-wise computations or reconstruct intermediate feature maps~\cite{yan2020cache,yuan2024survey}.
Such attacks exploit the predictable behavior of neural computations, specially during convolution and activation phases allowing adversaries to estimate gradients or extract partial model parameters without direct access to memory.

To mitigate these risks, several secure model execution techniques have been proposed. Lightweight defenses include noise injection, data perturbation, and DP-based randomization~\cite{xiao2019edge,ansari2020security,wang2021enabling,van2019tale}, which obscure timing and power signatures. Experimental evaluations report only 2.1\% power overhead and an additional 0.62 ns path delay with these methods~\cite{bhade2024lightweight}, making them practical for low-power System-on-Chip (SoC) designs. At the architectural level, strategies such as oblivious RAM (ORAM) and constant-access memory patterns~\cite{hua2018reverse} conceal access sequences, preventing attackers from mapping memory addresses to model parameters. Although these approaches significantly enhance confidentiality, they incur additional latency and energy consumption, limiting deployment in battery-constrained environments.

Beyond software-controlled randomization, layer-wise encryption and encrypted computation pipelines~\cite{moini2021power,wang2021enabling,wang2020mitigating,baciu2025secure} have proven effective for hiding intermediate data during inference.
These methods are often complemented by randomized memory access and hardware obfuscation to disrupt repetitive access patterns.
However, synchronization challenges arise when such techniques are applied to distributed micro-edge hardware, potentially reducing overall throughput.
To address these performance-security trade-offs, researchers have proposed compression based (quantization/pruning) perturbation masking \cite{liu2025quantization,yuan2022membership, yu2023x} and dynamic model encryption that adaptively modify the protection level according to resource availability.

Recent advancements have extended TEE capabilities to support secure model partitioning and verifiable inference. Systems such as Slalom~\cite{tramerslalom}, ShadowNet~\cite{sun2023shadownet}, and TEE-Shielded DNN Partition (TSDP)~\cite{zhang2024no} demonstrate how enclaves can execute sensitive model layers while delegating heavy computation to untrusted accelerators.
These hybrid designs yield 6-20$\times$ throughput improvements for secure inference compared to full-software solutions but introduce approximately 40\% transmission delay due to encrypted offloading overheads.
Furthermore, the Lyapunov-based adaptive privacy framework~\cite{cheng2025privacy} dynamically balances latency and privacy by adjusting partition frequency and task allocation, ensuring predictable privacy loss within real-time constraints. Nonetheless, intermediate feature exposure remains a concern when partially processed tensors traverse untrusted hardware, creating potential leakage channels.

Overall, secure model execution in onboard learning requires a careful integration of hardware trust anchors, stochastic obfuscation, and adaptive encryption. While TEEs and DP-based mechanisms form a strong defensive foundation, residual vulnerabilities persist in cache behavior, power analysis, and communication channels. Future research must focus on cross-layer co-design, combining hardware-aware adversarial defenses with context-sensitive encryption to achieve verifiable, energy-efficient, and attack-resilient learning at the edge.

\subsection{Privacy-Preserving Inference}

Ensuring privacy during inference is as critical as securing training, particularly in applications such as healthcare, finance, and autonomous systems. Privacy-preserving inference enables models to process encrypted or partitioned data without exposing sensitive inputs. Cryptographic methods like Fully Homomorphic Encryption (FHE) and Secure Multi-Party Computation (SMPC) offer strong confidentiality guarantees by performing computations directly on encrypted data~\cite{gross2020hardware,vedadi2023efficient,dai2023force}. However, their substantial computational and memory demands restrict feasibility on lightweight edge devices.

To overcome these limitations, hybrid frameworks such as ShadowNet~\cite{sun2023shadownet} and MirrorNet~\cite{liu2023mirrornet} combine TEE-secured execution with selective offloading to untrusted accelerators. By partitioning model layers between trusted and untrusted domains, they maintain privacy while reducing inference latency and energy consumption. These systems bridge cryptographic protection with hardware-based isolation, enabling near real-time secure inference at the edge.

Robustness remains equally vital. Adversarial perturbations (e.g., FGSM, DeepFool, AutoAttack)~\cite{ozbulak2020perturbation} can mislead models even under encryption. Defensive schemes such as Attention-Guided Knowledge Distillation (AGKD) and Bidirectional Metric Learning (BML)~\cite{wang2021agkd,mehta2024layered} improve resilience, recovering accuracy from 59.97\% to 99.18\%, albeit with high computational cost. Integrity threats like update poisoning~\cite{zhu2023research,hitaj2017deep,zhang2020poisongan} and model extraction~\cite{rakin2022deepsteal,gong2021model} are mitigated through DP-based federated learning and encrypted model blobs, though they may reduce accuracy by up to 24\%. Moreover, lightweight defenses such as X-Pruner~\cite{yu2023x} integrate randomization during pruning to minimize leakage, offering an efficient balance between privacy, robustness, and edge feasibility.

\subsection{Ensuring Trust and Transparency: Explainable AI}

As deep learning models increasingly operate in decentralized and autonomous edge environments, ensuring transparency and trust has become a fundamental requirement. However, the inherently opaque nature of deep neural networks (DNNs) complicates the interpretation of their decision-making processes, leading to concerns regarding fairness, accountability, and adversarial robustness~\cite{liang2021explaining,chen2019deep}. Unlike centralized AI systems that benefit from extensive computational resources for interpretability, onboard and edge learning must rely on lightweight explainability mechanisms capable of operating in real time without compromising efficiency or privacy.

Conventional Explainable AI (XAI) methods such as SHAP and LIME provide valuable interpretability but impose considerable computational and memory overheads, rendering them impractical for edge devices~\cite{shrotri2022constraint}. Recent innovations in approximate Shapley value estimation and model-agnostic interpretability~\cite{boselli2025explainable} have made low-cost XAI integration feasible in constrained environments. In parallel, Federated Explainable AI (FED-XAI) extends federated learning by embedding interpretability constraints within the training process, enabling models to generate privacy-preserving, human-interpretable insights directly on local nodes~\cite{bechini2023application,zhang2024survey}.

However, emerging research has revealed that XAI itself can introduce new privacy and security risks. Specifically, studies show that around 22\% to 50\% of encrypted network traffic can be classified using explanation-based leakage~\cite{senevirathna2025enhancing}, demonstrating that interpretability outputs such as gradients or saliency maps can inadvertently expose private information. This challenge is particularly acute for micro-edge devices, where limited computational resources restrict the ability to randomize or obfuscate explanation data, increasing the risk of explanation leakage and interpretability-based inference attacks.

Beyond transparency, XAI contributes directly to adversarial robustness. Embedding explainability into pruning and optimization pipelines allows models to dynamically remove redundant or high-risk parameters, enhancing both computational efficiency and resilience to adversarial perturbations~\cite{cantini2024xai,yu2023x}. Security-aware pruning further ensures that only the most informative and stable features influence model decisions, reducing the likelihood of model drift, gradient masking, and backdoor exploitation~\cite{mahfuz2024x,villar2023edge}. Furthermore, explainability-driven adversarial detection mechanisms enable onboard AI systems to identify, interpret, and respond to attacks in real time, significantly improving operational trust in safety-critical environments. 
In summary, integrating explainability into security-aware edge learning bridges the gap between transparency, efficiency, and trust. Lightweight interpretability techniques, federated XAI frameworks, and explanation-aware pruning collectively support privacy-preserving, robust, and interpretable AI, positioning onboard intelligence as a dependable solution for mission-critical and privacy-sensitive applications.

\textbf{Overview.} In onboard settings, security and privacy mechanisms do not simply wrap a learning pipeline; they reshape the feasible operating region for optimization and continual learning. Encryption, secure aggregation, and privacy-preserving variants of federated and split learning all add computation, memory, and communication overhead that directly competes with the same power and latency budgets targeted by compression and inference optimization \cite{ansari2020security, hu2022membership, tan2025defending}. When models are aggressively compressed via pruning, low-bit quantization, or hardware-aware NAS payload sizes for parameters, activations, or gradients decrease, which lowers the cost of encrypting, transmitting, and storing sensitive information. At the same time, compressed models typically have reduced representational capacity and tighter accuracy margins, making them more sensitive to the additional noise, stochasticity, or protocol-induced delays introduced by differential privacy, secure aggregation, or relay-style protocols. In other words, the same compression that enables deployment on power-limited devices also reduces the tolerance of the model to privacy-induced perturbations, and this coupling must be taken into account when deciding how far compression can be pushed. The interaction with learning dynamics is equally strong. In decentralized or continual onboard learning, each additional privacy layer increases per-round cost and often reduces the effective information content of each update \cite{hu2022membership, tan2025defending}. To stay within energy and latency constraints, designers may respond by reducing update frequency, limiting the number of participating clients per round, or shrinking the subset of parameters that are allowed to change-choices that directly affect convergence speed, stability, and final model quality. Conversely, optimization techniques such as parameter-efficient tuning and split or partial model updates can confine privacy protection to a small fraction of the network, reducing cryptographic and communication overhead and making stronger privacy budgets feasible under the same resource envelope. From a system perspective, security and privacy thus form an additional axis in the accuracy–latency–energy design space: higher protection levels tend to push the system toward more communication, more computation, and slower learning, while stronger compression and more efficient update mechanisms pull in the opposite direction. Effective onboard deployment of deep learning therefore, requires joint design of compression, inference policy, update strategy, and privacy mechanisms, rather than treating security as a post hoc layer added on top of an otherwise fixed optimization and learning pipeline.

\section{Advanced Topics}
\label{advanced_topics}

Onboard learning continues to evolve, requiring advanced techniques to balance efficiency, adaptability, and robustness in resource-constrained environments. This section explores three critical areas: (1) bridging model compression with continual learning to improve adaptability while preserving efficiency, (2) enhancing scalability and standardization to enable seamless deployment across heterogeneous edge platforms, and (3) leveraging hardware-software co-design to optimize AI for next-generation edge devices.

\subsection{Bridging Model Compression with Continual Learning}

Continual Learning (CL) enables AI models to incrementally learn new tasks without catastrophic forgetting, making it crucial for long-term deployment in dynamic environments. However, CL is constrained by high memory requirements, computational overhead, and knowledge retention challenges that are especially critical for onboard AI. Model compression techniques such as pruning, knowledge distillation, and quantization offer solutions by optimizing memory usage and inference speed while preserving task knowledge.

\subsubsection{Pruning for Continual Learning.} Integrating pruning into CL helps onboard models adapt efficiently while maintaining a compact structure. Sparse continual learning (SparCL)~\cite{wang2022sparcl} employs weight sparsity~\cite{kurtz2020inducing}, gradient sparsity~\cite{evci2022gradient}, and dynamic masking to selectively retain critical weights and prune redundant ones, minimizing memory footprint. Task-aware dynamic masking (TDM) further enhances retention by ensuring that essential parameters are preserved. Alternative methods such as gradient pruning~\cite{hung2019compacting} and continual prune-and-select (CP\&S)~\cite{dekhovich2023continual} further improve memory allocation by structuring models into reusable subnetworks optimized for incremental learning.

\subsubsection{Knowledge Distillation for Continual Learning.} Traditional experience replay is often infeasible for onboard AI due to memory constraints and privacy concerns. Knowledge Distillation (KD) mitigates catastrophic forgetting by transferring past knowledge to new models without requiring explicit data storage~\cite{li2017learning, zhang2020class}. Learning without Forgetting (LwF)~\cite{li2017learning} preserves past knowledge by constraining parameter updates, while deep model consolidation~\cite{zhang2020class} further refines inter-task representations. Prototype-sample relation distillation~\cite{asadi2023prototype} and batch-level distillation~\cite{fini2020online} improve adaptation in non-stationary environments by optimizing feature transfer between tasks.

\subsubsection{Quantization for Continual Learning.} Quantization reduces model size while preserving past knowledge, ensuring efficient adaptation in resource-limited environments. Adaptive quantization modules (AQM)~\cite{caccia2020online} dynamically adjust compression rates based on data complexity, while Bit-Level Information Preserving (BLIP)~\cite{shi2021continual} applies weight quantization to maintain knowledge retention. Vector Quantization Prompting (VQ-Prompt)~\cite{jiao2024vector} enhances continual learning by improving task abstraction and reducing memory footprint.

By integrating pruning, knowledge distillation, and quantization, continual learning models can efficiently adapt to new tasks while operating within strict resource constraints, enhancing onboard AI scalability and robustness.

\subsection{Scalability and Standardization in Onboard AI}

The scalability and reproducibility of onboard learning systems fundamentally depend on the capacity to evaluate AI models consistently across heterogeneous hardware. As onboard and on-device intelligence becomes central to edge computing, standardized benchmarking protocols have become indispensable for measuring performance, latency, and energy efficiency in a reproducible and comparable manner across System-on-Chip (SoC) architectures. Scalability ensures that models maintain functional integrity across diverse resource configurations, while standardization provides the methodological foundation for cross-platform reproducibility and fair performance evaluation.

\subsubsection{Scalability through Model Optimization.} Onboard AI must adapt to limited computational budgets without compromising prediction accuracy or responsiveness. Model compression methods, including pruning, quantization, and distillation (as reviewed in Section~\ref{sec:model_compression}), mitigate resource constraints by reducing inference complexity~\cite{mayer2020scalable, tyagi2022scadles}. However, modern deployment scenarios increasingly demand adaptive optimization across heterogeneous SoCs. Hardware-Aware Neural Architecture Search (HW-NAS) frameworks address this by coupling architecture exploration with device-specific latency, power, and thermal models~\cite{luo2022lightnas, chitty2022neural}. Such frameworks automatically identify Pareto-efficient architectures under multi-objective constraints, ensuring that onboard models scale seamlessly across CPU-GPU-NPU configurations while maintaining predictable latency and energy behavior.

%======================================================

% \Figure[t!](topskip=0pt, botskip=0pt, midskip=0pt)[width=0.7\textwidth]{Fig_5_codesign.pdf}
% { \textbf{integrating neural architecture search, hardware exploration, and software optimization into a unified workflow. Model-level design adapts network topology to hardware constraints, hardware-level customization enhances computation and memory efficiency, and softwarelevel deployment refines kernels and memory management for real-time onboard inference.}\label{codesign}}

%======================================================

\subsubsection{Standardization for Onboard AI Deployment.} Standardized model representation and execution pipelines are crucial to achieving platform-agnostic scalability. Frameworks such as the Open Neural Network Exchange (ONNX) standardize model serialization across backends like TensorRT, OpenVINO, and CoreML~\cite{zhou2024investigating}, while the Multi-Level Intermediate Representation (MLIR) provides a modular compiler layer that unifies hardware-specific code generation and operator optimization across diverse accelerators~\cite{lattner2021mlir, majumder2023hir}. These abstractions ensure functional equivalence across hardware targets, reducing divergence between training and deployment environments. Benchmarking frameworks such as MLPerf Tiny extend this philosophy by introducing standardized workloads, fixed evaluation seeds, and power-latency normalization protocols for resource-limited devices~\cite{akkisetty2025overview, tabanelli2023dnn}. Together, they establish a baseline for reproducibility and interoperability in onboard and edge AI.

\subsubsection{Standardized Energy and Latency Benchmarking Across Heterogeneous Edge Platforms}

Reliable evaluation of latency and energy consumption across heterogeneous edge platforms is essential for assessing onboard and edge-AI performance. Modern deployments span both heterogeneous System-on-Chip (SoC) architectures integrating CPU, GPU, and NPU units, as well as fundamentally different platforms including microcontrollers, embedded GPUs, and mobile processors. Variability in clock scaling, memory hierarchies, and power instrumentation complicates comparison when benchmarking procedures are not harmonized, even for nominally identical workloads~\cite{varghese2021survey}. Cross-platform evaluation is particularly challenging as microcontrollers (10--250\,MHz Cortex-M cores), embedded GPUs (e.g., NVIDIA Jetson at GHz scales), and mobile SoCs with proprietary NPUs expose distinct execution and measurement constraints~\cite{varghese2021survey}.

MLPerf Tiny, developed by MLCommons, represents an important step toward cross-platform standardization by defining fixed inference tasks and prescribed measurement procedures for latency and energy~\cite{banbury2021mlperf}. However, its applicability to adaptive and on-device learning remains limited. Separate configurations are required for latency and energy measurement, preventing joint evaluation under identical operating conditions~\cite{bartoli2025benchmarking}. In addition, single-trigger synchronization conflates inference with auxiliary operations such as model loading, activation transfers, and gradient computation, which dominate memory traffic during on-device learning but remain indistinguishable in aggregate measurements~\cite{bartoli2025benchmarking}. Hardware constraints further complicate energy measurement, as external monitors often operate outside the voltage ranges required by advanced SoC cores, introducing unmeasured overhead through additional regulation stages~\cite{bartoli2025benchmarking}.

The impact of phase-agnostic benchmarking on cross-platform comparability has been quantified by Bartoli et al.~\cite{bartoli2025benchmarking}, who introduced phase-separated measurement using dual-trigger synchronization and high-precision shunt instrumentation on the STM32N6 platform. Across 1000 runs per MLPerf Tiny workload, standardized timing and power synchronization reduced the relative Energy-Delay Product by 21-28\% across DS-CNN, MobileNet, ResNet-8, and autoencoder models. Phase separation further revealed that pre- and post-inference operations accounted for the majority of energy consumption in DS-CNN, contributions that conventional protocols attribute uniformly to inference~\cite{bartoli2025benchmarking}.

These limitations directly affect the evaluation of parameter-efficient fine-tuning methods deployed across heterogeneous edge platforms. Techniques such as TinyTL and LoRA reduce training-time computation and parameter updates but do not eliminate data movement costs associated with model loading, synchronization, and aggregation, which remain conflated with inference under existing benchmarking practices~\cite{cai2020tinytl, hu2022lora}. Architectural diversity further obscures interpretation, as identical configurations exhibit divergent latency and memory behavior across microcontroller SRAM, embedded GPU DRAM, and mobile NPU on-chip memory systems~\cite{varghese2021survey}.

While benchmarking suites such as MLPerf Inference Edge and ELAC strengthen cross-device inference comparability through controlled runtime configurations, they remain inference-centric and omit training-specific costs such as gradient computation, update bandwidth, and communication overhead~\cite{bartoli2025benchmarking}. As edge intelligence increasingly incorporates adaptive and on-device learning, standardized benchmarking must extend beyond inference to support phase-separated, training-aware, and device-agnostic energy and latency measurement across heterogeneous platforms.

\subsection{Hardware Co-Design for Next-Generation onboard learning}

To meet the challenging performance and energy demand of next generation onboard AI, it requires hardware-software co-design approach for optimized learning and inference execution. Traditional hardware\& computing architectures, designed for general purpose applications, often fail to meet the computational and latency requirements of deep learning workloads, primarily due to data movement inefficiencies and limited memory bandwidth. These challenges have driven the advancement of customized accelerators and co-optimized learning frameworks that unify algorithmic design with hardware execution to deliver real-time intelligence in energy-constrained environments.

\Figure[t!](topskip=0pt, botskip=0pt, midskip=0pt)[width=0.8\textwidth]{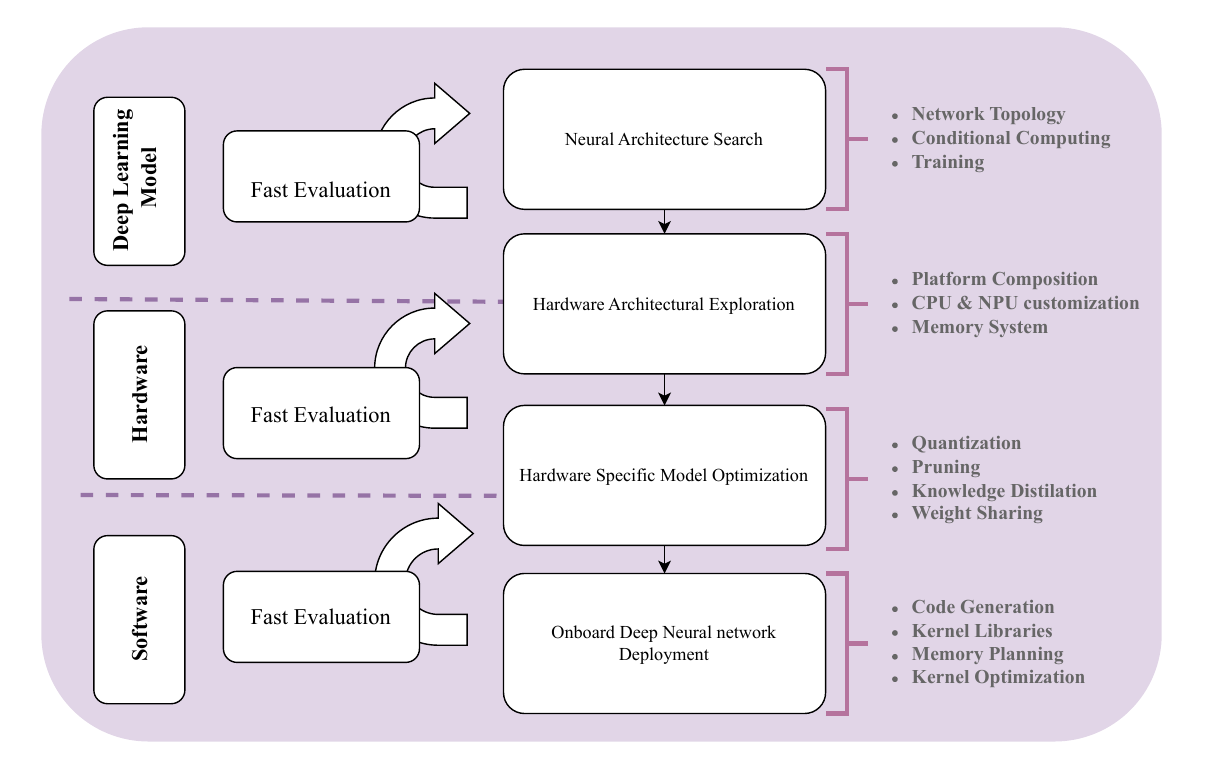}
{%
  \begin{minipage}[t]{0.8\textwidth} % <-- [t] makes the box top-aligned
    \raggedright
    \textbf{Integrating neural architecture search, hardware exploration, and software optimization into a unified workflow Model-level design adapts network topology to hardware constraints, hardware-level customization enhances computation and memory efficiency, and software level deployment refines kernels and memory management for real-time onboard inference.}
  \end{minipage}%
  \label{codesign}%
}

\subsubsection{Application-Specific Hardware for Onboard AI.} 

Among the most significant advancements enabling onboard intelligence are Application-Specific Integrated Circuits (ASICs) and System-on-Chip (SoC) architectures designed for AI workloads~\cite{lee2020hardware}. Unlike general purpose processors, these devices integrate heterogeneous compute elements through CPU-GPU-NPU, and accelerators on a single platform to maximize parallelism and minimize data movement overhead. This architectural specialization provides substantial gains in power efficiency and throughput density, key for embedded and edge devices operating within strict energy budgets. The introduction of Compute-in-Memory (CIM) and dataflow based processing architectures further mitigates the traditional memory bottleneck by co-locating computation and data storage, drastically reducing energy wasted on data transfers \cite{diao2025computing}. This configurations also enable low latency inference and incremental learning directly onboard eliminating the need for continuous cloud connectivity.

\subsubsection{Holistic Model-Hardware Co-Design.} 
While application-specific hardware offers efficiency gains, true onboard intelligence demands co-optimization between model architectures and underlying hardware platforms. Holistic model hardware co-design framework aligns algorithmic structure with architectural characteristics, ensuring that neural networks are not only compact but also hardware executable. These algorithmic adaptations are designed to harmonize directly with hardware acceleration mechanisms, ensuring that both layers evolve cooperatively rather than independently. The general holistic architecture of hardware co-design for onboard learning, illustrated in Figure~\ref{codesign}, encapsulates this multi-level synergy as an integrated deployment framework. At the model level, topology-aware NAS refines network structures to align with hardware constraints, while techniques such as quantization, pruning, and weight sharing optimize computational efficiency~\cite{zhang2022algorithm}. The hardware level focuses on customized CPU/NPU acceleration, enhancing efficiency by optimizing platform composition and memory access. Meanwhile, the software level streamlines onboard DNN deployment through optimized code generation, kernel libraries, and memory management. By integrating hardware-aware AI model design, onboard learning can achieve scalable, high-performance deployment across next-generation edge devices.

\subsubsection{Cloud-Edge-Device Co-Inference.}
 
Running full deep models locally often results in high latency and performance trade-offs, whereas offloading inference entirely to the cloud introduces dependency on connectivity and raises data privacy concerns. 
To address this limitation, edge-cloud co-inference has emerged as a cooperative deployment paradigm that partitions deep networks between the device and the cloud, enabling early layers to execute locally for rapid feature extraction while delegating deeper layers to high-performance servers for complex processing. 
This approach reduces resource bottlenecks but also requires coordinated optimization of model architecture, feature encoding, and transmission bandwidth, forming a natural case of model hardware co-design.

Initial frameworks such as Neurosurgeon \cite{kang2017bottlenet} and DeepThings \cite{zhang2018deepsplit} demonstrated workload partitioning to reduce end-to-end latency and improve system responsiveness. SplitInfer \cite{teerapittayanon2017distributed} extended this concept through dynamic layer scheduling under variable network conditions. Furthermore, Hassan et al. \cite{hassan2025spikebottlenet} introduced SpikeBottleNet, a spike-driven feature compression architecture that integrates spiking activations with encoder-decoder bottleneck modules in convolutional networks. Experimental evaluations showed up to 256-fold bit compression in the final convolutional layer of ResNet with less than 0.2\% accuracy loss, and an energy efficiency gain exceeding 140$\times$ compared with the baseline BottleNet. These findings highlight how spike-based, event-driven processing can co-optimize communication and computation, enabling scalable, low-power onboard deployment and adaptive edge-cloud learning.
Zhou et al. \cite{zhou2024graph} further advanced the concept through GCoDE, which co-optimizes Graph Neural Network architecture and device-edge mapping, realizing up to 44.9$\times$ speedup and 98.2\% energy savings. 
Similarly, model-architecture co-design for temporal GNN inference on FPGA is presented in \cite{zhou2022model} , achieving up to 8.8$\times$ higher throughput and over 5$\times$ lower latency through joint algorithmic and hardware optimization.  

In addition to software-hardware co-design efforts, streaming and continual inference approaches allow onboard models to adjust progressively to new data without requiring complete retraining. These methods update model parameters on the fly, maintaining accuracy as operational conditions change while keeping computational cost low. Such adaptive inference extends co-design concepts toward lifelong learning, supporting reliable and efficient performance during extended or safety-critical edge missions \cite{banerjee2023streaming}.

\section{Discussion}

\subsection{Unification Across Pillars}
The survey conducted across Sections~II–VI indicates that current research on edge intelligence remains dominated by pillar-specific optimization, despite the fact that deployment behavior is governed by interactions among these components. While advances in model compression, inference efficiency, decentralized learning, security, and deployment frameworks have each produced measurable gains, their effectiveness in operational environments depends on how design decisions across these areas interact under shared constraints. In resource-constrained edge systems, performance and reliability are therefore determined by the joint configuration of these elements rather than by the isolated strength of any single technique.

This interdependence is particularly evident in model compression. Approaches such as pruning, quantization, and structured sparsification are commonly assessed using static accuracy and model size metrics. However, these techniques also alter memory access patterns, operator granularity, numerical precision, and gradient behavior, which directly influence runtime scheduling, inference latency, and learning stability on heterogeneous hardware platforms~\cite{wang2019haq, xia2022structured, frantar2023sparsegpt, hoefler2021sparsity, gross2020hardware}. As a result, compression strategies that appear favorable under offline evaluation may lead to degraded or unstable system behavior when their interaction with inference pipelines and hardware execution paths is not explicitly considered.

A similar coupling arises between inference optimization and online or decentralized learning. In edge environments, inference execution and model updates compete for limited compute, memory, and energy resources. Decisions regarding update frequency, communication volume, and synchronization therefore have direct consequences for real-time inference performance~\cite{li2020federated, imteaj2021survey, varghese2021survey}. Parameter-efficient adaptation methods reduce the cost of updates, but their practical benefit depends on runtime support for selective parameter activation and memory-aware execution, which are themselves deployment-level considerations~\cite{cai2020tinytl, wang2025loraedge, zhao2025hardware}. Learning behavior cannot be meaningfully evaluated without reference to the inference and execution context in which it operates.

Security and privacy further constrain system design in ways that resist modular treatment. Protection mechanisms operate on intermediate representations and update signals whose structure is determined by model architecture, compression strategy, and learning protocol. Their effectiveness, overhead, and impact on convergence therefore vary across configurations and cannot be treated as independent safeguards~\cite{ansari2020security, zhao2020idlg, hu2022membership, tan2025defending}. In edge settings with strict resource limits, privacy guarantees are only meaningful when they are designed in coordination with model structure and update dynamics.

These observations point to the need for a shift away from sequential pipelines toward system-level co-design. Compression, inference, learning, and security should be formulated as jointly constrained variables within optimization frameworks that explicitly account for latency, energy consumption, communication limits, and robustness requirements~\cite{chitty2022neural, luo2022lightnas, yu2020easiedge}. Such an approach reframes optimality in terms of stable and reproducible system behavior under realistic operating conditions, rather than isolated improvements measured under controlled assumptions. This perspective provides the conceptual basis for the subsequent analysis of deployment risks and operational considerations.

\subsection{Lessons Learned}

A primary lesson emerging from this survey is that the effectiveness of onboard learning systems is determined less by individual algorithmic advances than by the coherence with which optimization, inference, learning, privacy, and hardware constraints are integrated at the system level. The extensive body of work on model compression including pruning, quantization, knowledge distillation, neural architecture search, and parameter efficient fine-tuning demonstrates that deep learning models can be made deployable on resource-constrained platforms without prohibitive loss in task performance when compression strategies are aligned with hardware characteristics \cite{wang2019haq, cai2020tinytl, tanaka2020pruning, frantar2023sparsegpt, xia2022structured, zhang2024oncenas, itahara2021distillation, wang2025loraedge}. However, these gains are predominantly presented under static deployment assumptions, revealing a persistent disconnect between offline optimization objectives and the dynamic operational realities of onboard systems.

A second lesson concerns inference efficiency. Techniques such as early-exit inference, adaptive computation, and task-aware execution have proven effective in reducing latency and energy consumption, particularly under tight real-time constraints \cite{wang2019adaptive, yu2020easiedge, sun2023shadownet}. Yet, inference optimization is frequently treated as a post hoc enhancement, decoupled from model design and learning strategy. This separation limits the ability of onboard systems to adapt inference behavior in response to fluctuating resource availability, communication conditions, and mission priorities.

The survey further reveals that decentralized and federated learning frameworks are indispensable for enabling onboard learning under limited connectivity and data-sharing constraints \cite{li2020federated, imteaj2021survey, varghese2021survey, li2023predictive}. Despite this progress, continual onboard learning remains comparatively underdeveloped. Existing methods struggle to balance adaptability, memory overhead, energy consumption, and long-term stability under non-stationary task distributions \cite{ro2021autolr, li2025improving, senevirathna2025enhancing}. This indicates that learning over time cannot be treated as an incremental extension of cloud-based training paradigms; instead, it must be formulated as a resource-constrained optimization and control problem intrinsic to onboard operation.

From a security and privacy perspective, the surveyed literature confirms both the necessity and the cost of protection mechanisms. Techniques such as secure aggregation, gradient leakage mitigation, and privacy-preserving learning reduce information exposure but introduce measurable overheads in computation, communication latency, and accuracy \cite{ansari2020security, zhao2020idlg, hu2022membership, tan2025defending}. A key lesson is that privacy and security are not auxiliary concerns but first-order design constraints that must be co-optimized with learning and inference under strict onboard resource budgets.

Finally, insights from hardware-aware and cross-layer design studies underscore that sustainable onboard intelligence requires tighter integration between algorithms, software runtimes, and hardware architectures \cite{chitty2022neural, hoefler2021sparsity, gross2020hardware, zhao2025hardware}. The absence of standardized, system-level benchmarks across heterogeneous platforms remains a significant barrier to reproducibility, comparability, and principled design.

\subsubsection{Edge Deployment}

Edge or onboard deployment provides the context in which interactions among compression, inference, learning, and security are ultimately resolved. In deployed systems, algorithmic advances are mediated by system interfaces, including model representation, compilation, runtime scheduling, and measurement infrastructure. Standardized model formats and compiler-based toolchains are therefore essential for enabling portability and reproducibility across heterogeneous edge platforms~\cite{zhou2024investigating, lattner2021mlir, majumder2023hir}. Without such support, optimizations such as sparsity and quantization frequently fail to translate into consistent performance gains due to mismatched kernels and memory layouts~\cite{hoefler2021sparsity, frantar2023sparsegpt, gross2020hardware}.

Deployment constraints also motivate alternative execution paradigms. When full on-device inference is infeasible, split execution and edge–cloud co-inference enable computation to be partitioned in response to resource limitations and communication conditions~\cite{kang2017bottlenet, teerapittayanon2017distributed, zhang2018deepsplit}. Partitioning decisions interact with compression strategies, feature representations, and privacy exposure, and must therefore be optimized jointly rather than treated as architectural afterthoughts~\cite{hassan2025spikebottlenet}. Similarly, decentralized learning at the edge requires coordination between inference workloads and update schedules. Parameter-efficient updates become effective only when integrated into runtime policies that manage resource allocation across tasks~\cite{li2020federated, imteaj2021survey, varghese2021survey, li2023predictive, wang2025loraedge}.

Reliable edge or onboard deployment further depends on careful measurement and reproducibility. Energy consumption, latency, and stability are sensitive to hardware configuration and runtime behavior on modern edge systems. While deployment frameworks enable these integrations in practice, they also expose systems to operational variability that is not captured by controlled evaluations. Examining how deployed systems fail under such conditions is therefore essential for understanding the limits of current approaches.

\subsubsection{Real-World Failure Modes and Deployment Risks}
The consequences of insufficient unification become apparent when edge intelligence systems are examined under real deployment conditions. Many approaches surveyed in this work demonstrate strong performance under controlled benchmarks; however, they exhibit fragility when exposed to the variability inherent in operational environments. One prominent source of failure is resource non-stationarity. Edge platforms frequently experience fluctuations in available compute and memory due to thermal constraints, power management, background workloads, and intermittent connectivity. Methods optimized under fixed-budget assumptions may therefore violate latency or energy constraints once deployed, particularly when inference and learning schedules are managed independently~\cite{wang2019adaptive, yu2020easiedge, sun2023shadownet}.

Furthermore, failure mode arises from long-term distributional drift. In dynamic environments, models must adapt continuously while operating within tight resource envelopes. However, continual and online learning mechanisms remain sensitive to aggressive compression and quantization, which can reduce representational capacity and alter optimization dynamics over time~\cite{ro2021autolr, li2025improving, senevirathna2025enhancing}. These effects are rarely captured by short-horizon evaluations, yet they dominate system behavior during prolonged deployment.

Security and privacy introduce additional risks. Although decentralized learning limits direct data sharing, intermediate updates and representations can leak sensitive information or enable inference attacks under realistic threat models~\cite{zhao2020idlg, hu2022membership}. Defensive mechanisms impose computational and communication overheads and may interfere with convergence if not co-designed with the underlying model and protocol~\cite{tan2025defending, ansari2020security}. Furthermore, discrepancies between training environments and deployment runtimes, including differences in compilers, kernels, and quantization behavior, can produce silent performance regressions that undermine reliability and reproducibility~\cite{zhou2024investigating, lattner2021mlir, majumder2023hir, bartoli2025benchmarking}.

These failure modes indicate that fragmentation across methodological pillars is not merely inefficient but can compromise system reliability. Addressing these risks requires evaluation methodologies that explicitly account for resource variability, temporal non-stationarity, and adversarial exposure, rather than relying solely on static accuracy and efficiency metrics.

\subsection{Future Outlook}

Looking ahead, the next phase of research in onboard optimization and learning will require a shift from technique-centric improvements toward holistic, decision-oriented system design. One promising direction lies in multi-objective and hardware-aware optimization frameworks that jointly consider accuracy, latency, and energy throughout the model lifecycle, rather than optimizing these dimensions in isolation. While neural architecture search and compression-aware training provide an initial foundation, further work is needed to improve portability, reduce design-time cost, and support deployment across heterogeneous platforms.

Another critical research direction concerns continual and adaptive onboard learning. Future systems must support learning under non-stationary conditions while respecting strict memory and energy constraints. This calls for resource-aware continual learning frameworks in which model capacity, update frequency, and learning objectives are dynamically adjusted based on real-time resource feedback, extending existing work on adaptive learning rates and parameter-efficient adaptation \cite{ro2021autolr, wang2025loraedge}. Such approaches are essential for enabling long-term autonomy in dynamic environments.

Privacy-aware onboard learning represents an additional open challenge. Rather than imposing static privacy guarantees, future systems should explore adaptive protection mechanisms that explicitly trade privacy, accuracy, and efficiency in response to mission context and resource availability. Co-designing privacy mechanisms with optimization and learning strategies will be particularly important for autonomous systems operating in untrusted or adversarial environments.

More broadly, advances in cross-layer co-design are likely to play a central role in the evolution of onboard intelligence. Closer integration between algorithms, runtime systems, and emerging hardware accelerators offers a pathway toward scalable and dependable deployment. Establishing standardized evaluation methodologies that capture system-level trade-offs across heterogeneous platforms will be essential to bridge the gap between algorithmic innovation and real-world onboard operation.

In summary, future progress in onboard optimization and learning will depend on the development of unified system architectures in which compression, inference, learning, privacy, and hardware constraints are jointly and dynamically balanced. Achieving this integration is critical for enabling robust, adaptive, and energy-efficient onboard AI systems capable of autonomous operation in resource-limited and evolving environments.

\section{Conclusion}

This survey systematically examined methodologies for enabling efficient onboard learning across five key aspects: model compression, which reduces model size while maintaining accuracy, efficient inference, which minimizes latency and energy use, decentralized learning, which supports collaborative model adaptation without centralized dependence, security and privacy, which safeguard onboard intelligence from adversarial and data-related threats, and advanced topics that focus on cross-layer optimization through hardware-software co-design. Together, these aspects define the technical foundation for modern on-device deep learning and optimization.

Despite remarkable advances, onboard intelligence continues to face several challenges that shape its research frontier. Achieving continual learning in compact models remains difficult, as maintaining adaptability while conserving energy and memory often leads to performance degradation over time. Privacy-preserving methods still struggle to balance security with real-time responsiveness, particularly in environments that demand ultra-low latency. Integrating algorithms, compilers, and edge accelerators into cohesive co-design frameworks requires deeper synchronization to achieve scalable efficiency across diverse hardware platforms. The field also lacks standardized evaluation protocols that capture trade-offs between accuracy, energy, robustness, and latency, limiting cross-system reproducibility. Moreover, ensuring stability and self-adaptation in dynamic or adversarial conditions calls for new forms of autonomous optimization that can adjust without external retraining. Finally, as onboard systems become increasingly autonomous, embedding transparency, interpretability, and trustworthiness into their learning processes is essential for reliable deployment in safety-critical domains. By addressing these open challenges, future research can move onboard learning from efficient execution toward genuine edge intelligence systems that are adaptive, secure, and resilient, capable of operating independently in complex hardware-constrained environments.

\section*{Declaration of competing interest}

\noindent The authors declare that they have no known competing financial interests or personal relationships that could have appeared to influence the work reported in this paper.

\section*{Acknowledgments}

\noindent This work has been supported by the SmartSat CRC, whose activities are funded by the Australian Government’s CRC Program. This work use an open-source realistic satellite simulator (Basilisk and BSK-RL) that is actively developed by Dr. Hanspeter Schaub and team at AVS Laboratory, University of Colorado Boulder. Also, the authors would like to express their sincere gratitude to BAE Systems for their invaluable support and collaboration throughout this research.
% And this work utilized the GPU and computing resources at  Adelaide University

%\bibliographystyle{cas-model2-names}
%\bibliographystyle{IEEEtran}  
\bibliography{refs}

\begin{IEEEbiography}[{\includegraphics[width=1in,height=1.25in,clip,keepaspectratio]{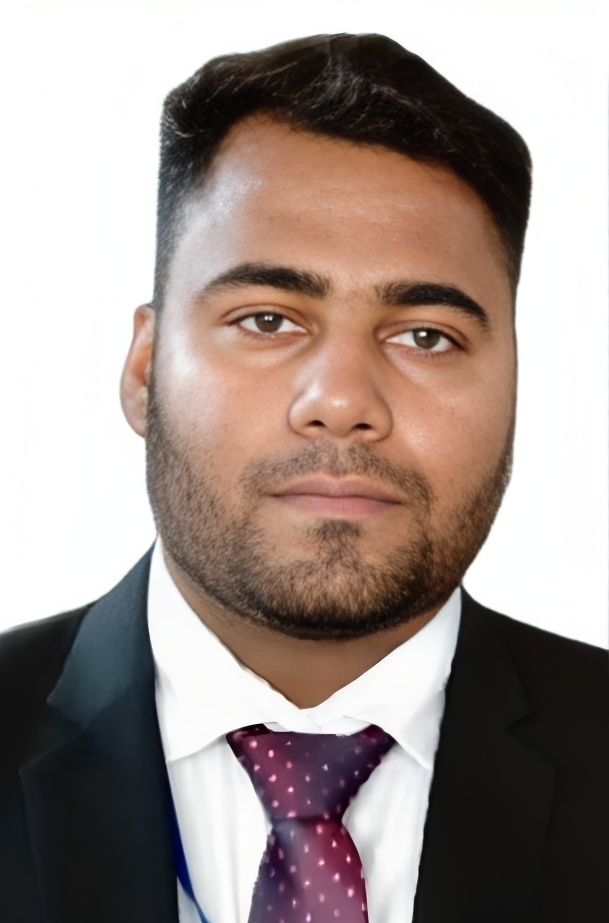}}]{Monirul Islam Pavel} is currently pursuing a Doctor of Philosophy (PhD) in Computer and Information Science at the  Adelaide University His research focuses on developing lightweight reinforcement learning algorithms for onboard implementation in satellite constellations, optimizing heterogeneous resources in dynamic environments. Pavel completed Master of Computing (by research) from the Faculty of Information Science and Technology (FTSM) at The National University of Malaysia, and Bachelor of Science in Computer Science and Engineering from BRAC University, Bangladesh. Following his graduation, he gained extensive experience working in the domain of AI across various industries in research and development sectors, as well as serving as research assistant.
\end{IEEEbiography}

\begin{IEEEbiography}[{\includegraphics[width=1in,height=1.25in,clip,keepaspectratio]{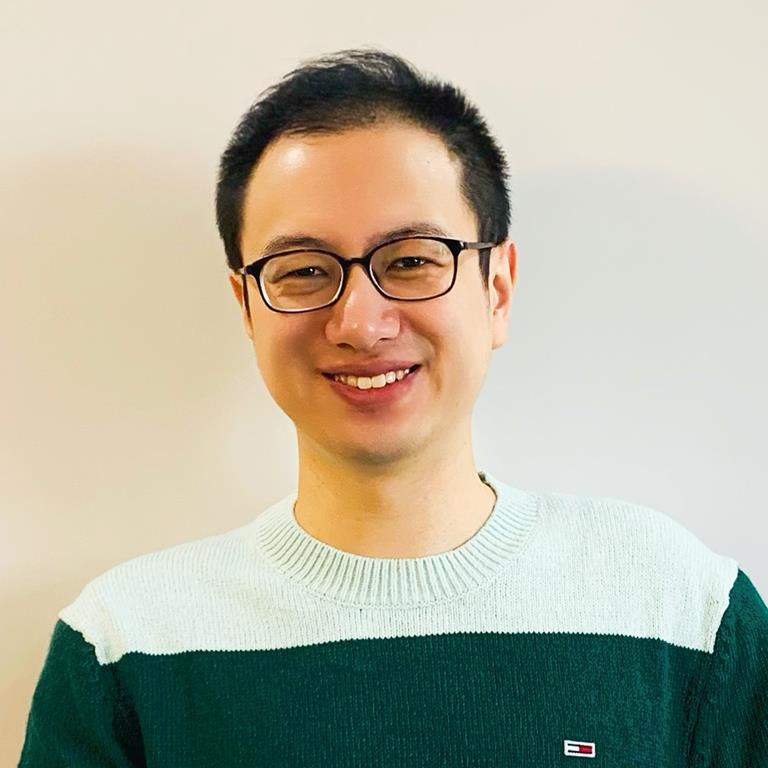}}]{Siyi Hu} is a Research Fellow in Artificial Intelligence at the University of South Australia, working within the SCARLET-$\alpha$ project on next-generation satellite autonomy. His research focuses on reinforcement learning, multi-agent systems, and onboard intelligence for space applications. He has published in top AI venues including ICLR, ICML, and AAAI, and contributes to open-source multi-agent learning libraries. He co-leads the design of reinforcement learning frameworks for resource-constrained satellite constellations in collaboration with SmartSat CRC and BAE Systems. Siyi holds a PhD in Computer Science from the University of Technology Sydney.
\end{IEEEbiography}

\begin{IEEEbiography}[{\includegraphics[width=1in,height=1.25in,clip,keepaspectratio]{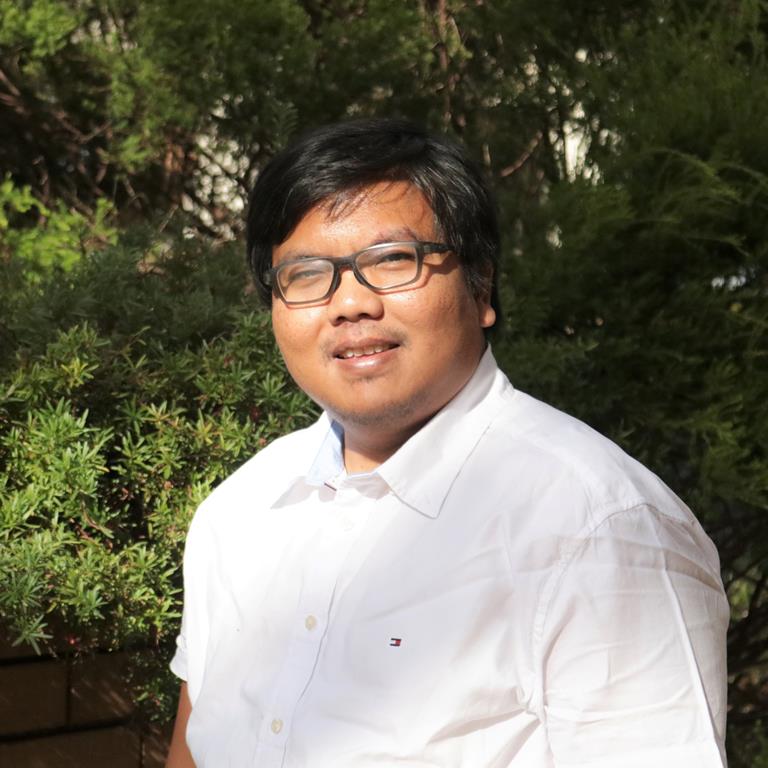}}]{ Mahardhika Pratama}  is currently an associate professor-level enterprise fellow in AI at the academic unit of STEM, University of South Australia, Adelaide, Australia. Prior to this appointment, he was an assistant professor at the school of CSE, Nanyang Technological University, Singapore from 2017 to Jan 2022, a lecturer at the department of CSIT, La Trobe University, Melbourne, Australia from 2015 to 2017 and a postdoctoral research fellow at the University of Technology Sydney, Sydney, Australia from 2014 to 2015. Mahardhika earned a PhD degree from the University of New South Wales, Australia in Nov 2014, a M.Sc degree from Nanyang Technological University, Singapore in Nov 2011 and a B.Eng degree from Institut Teknologi Sepuluh Nopember, Indonesia in Mar 2010. The research focus of Mahardhika is in deep learning where he is an active researcher in continual learning, lifelong learning, incremental learning, stream learning, fuzzy neural networks, and intelligent control systems. The goal of his study is to develop a deep learning algorithm that can learn forever as the case of biological learning. The application domain of his study includes manufacturing and satellite systems.
\end{IEEEbiography}

\begin{IEEEbiography}[{\includegraphics[width=1in,height=1.25in,clip,keepaspectratio]{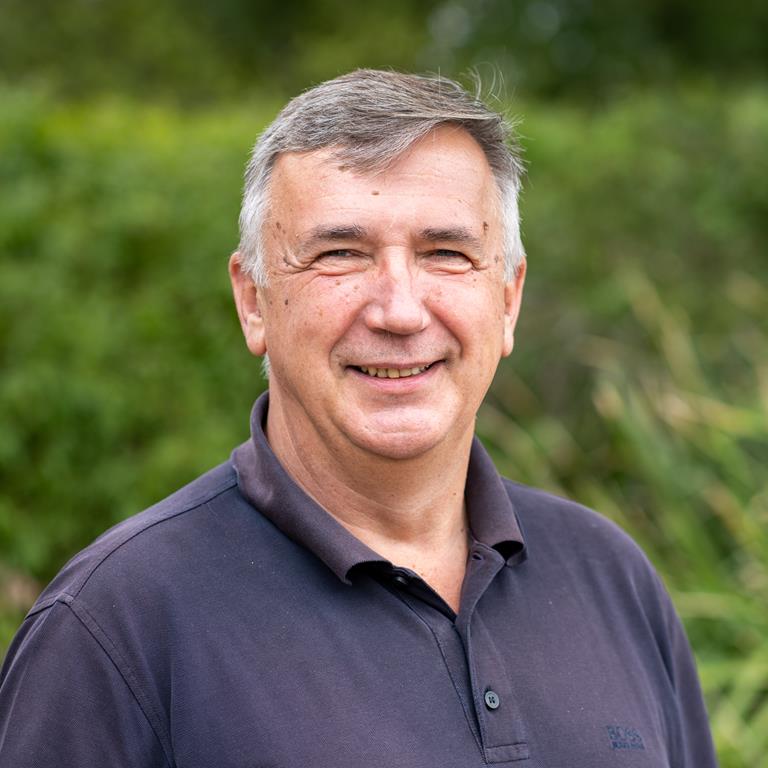}}]{Professor Ryszard Kowalczyk} joined the University of South Australia in 2022 from Swinburne University of Technology, where he was Director of Swinburne’s Key Lab for Intelligent Software Systems and Head of Distributed Artificial Intelligence (AI) Systems Research Group, and the former inaugural Wipro Chair of AI, also at Swinburne University of Technology in Melbourne. Prior to this, he led AI research at CSIRO and a number of corporate research and development centres in Australia and overseas. Professor Kowalczyk’s research interests include AI, autonomous software agents and collective intelligence for automated and optimised decision-making, and their applications in digital eco-systems of interacting systems, people and things, including cyber-physical-social eco-systems. It involves automated reasoning, negotiation, learning, adaptation and self-organisation in autonomous multi-agent systems, enabling smart infrastructure, smart cities, smart exchanges and industry 4.0. Professor Kowalczyk has co-authored four patents, published more than 250 scientific articles, and provided expert policy advice to government bodies in Australia and overseas.  Professor Kowalczyk received PhD in Computer Science (Artificial Intelligence) from Silesian University of Technology in Poland and holds a lifetime title of State Professor awarded by the President of Poland.
\end{IEEEbiography}

\EOD
\end{document}